%% file: main.tex
\documentclass[10pt,twocolumn,letterpaper]{article}
\usepackage{cvpr}
\definecolor{cvprblue}{rgb}{0.21,0.49,0.74}
\usepackage[pagebackref,breaklinks,colorlinks,allcolors=cvprblue]{hyperref}
\usepackage{multirow}
\usepackage{algorithm}
\usepackage{algorithmicx}
\usepackage{algpseudocode}
\usepackage{enumitem}
\usepackage{microtype}
\usepackage{pifont}
\usepackage{booktabs,tabularx}
\usepackage{listings}
\newcommand{\themethod}{EgoMirror\xspace}
\newcommand{\themethodfull}{\textls[0]{Generating Multi-View Robot Demonstrations from Egocentric Data\xspace}}
\title{\themethod: \themethodfull}

\author{
Hongwei Fan$^{1,2}$ \quad
Xinyu Wang$^{1,2}$ \quad
Chenbo Xin$^{1,2}$ \quad
Yuxuan Yao$^{1,2}$ \quad
Chengwei Zhang$^{1,2}$ \\
Yuxing Long$^{1,2}$ \quad
Jiafeng Xu$^{1}$ \quad
Xiaoqi Li$^{1,2}$ \quad
Hao Dong$^{1,2,\dagger}$ \\
$^{1}$CFCS, School of Computer Science, Peking University $^{2}$QYRO \\
{\tt\small hwfan25@stu.pku.edu.cn\quad \tt\small hao.dong@pku.edu.cn} \\
\textbf{\url{https://egomirror.github.io/}}
\vspace{-15pt}
} 

\begin{document}
\twocolumn[{
    \renewcommand\twocolumn[1][]{#1}
    \maketitle
    \vspace{-2em}
    \begin{center}
        \includegraphics[width=\textwidth]{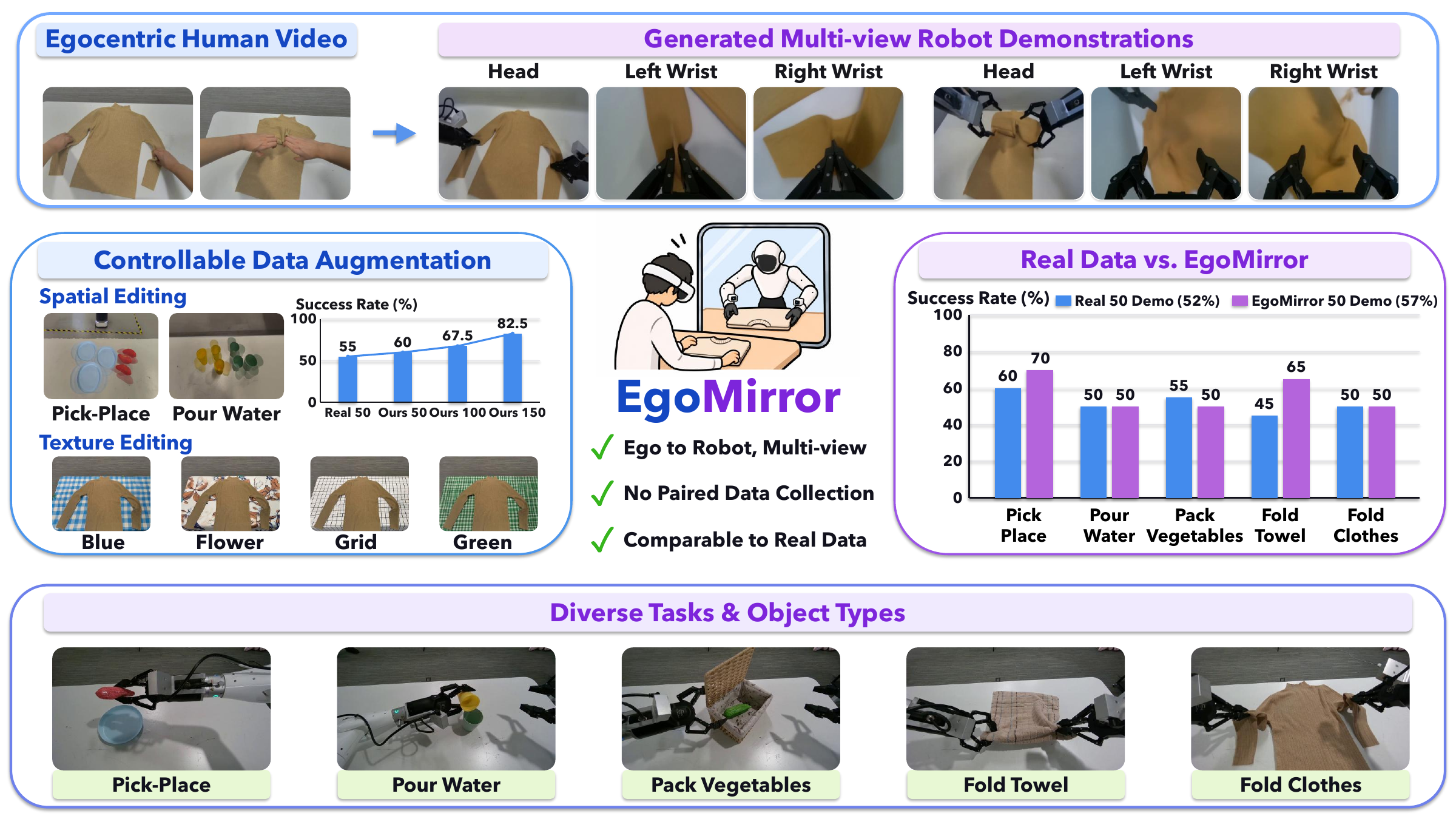}
        \captionof{figure}{
            Overview of \themethod. \themethod converts egocentric human videos into multi-view robot demonstrations without collecting human-robot paired data, achieving policy performance comparable to real-robot demonstrations across five tasks.
        }
        \label{fig:teaser}
    \end{center}
}]

\input{sec/0_abstract}

\input{sec/1_intro}
\input{sec/2_related_work}

\input{sec/3_method}
\input{sec/4_experiment}
\input{sec/5_conclusion}

\input{sec/llm_usage}

\input{sec/ack}
{
    \small
    \bibliographystyle{ieeenat_fullname}
    \bibliography{main}
}
\input{sec/X_suppl}

\end{document}

%% file: sec/0_abstract.tex
\begingroup
  \renewcommand{\thefootnote}{}
  \footnotetext{$\dagger$: Corresponding author.}
\endgroup

\begin{abstract}
\noindent General-purpose robot learning requires diverse demonstration data, yet collecting such data on physical robots is costly. Egocentric human videos offer a scalable alternative, but converting them into robot demonstrations remains challenging due to the observation gap between human egocentric views and multi-view robotic observations.
To bridge this gap, we propose \themethod, a framework that converts egocentric human demonstrations into multi-view robot demonstrations. Our multi-view video generation model jointly generates head and wrist videos conditioned on shared 4D scene geometry, robot renderings, and human hand poses. To train this model without collecting human-robot paired data, we construct paired supervision through reverse synthesis, retargeting robot gripper motions into diverse human hand poses and reconstructing 4D scenes. We further develop an end-to-end system that converts VR-collected human videos into multi-view robot demonstrations with robot actions.
Experiments on five real-world manipulation tasks show that policies trained on 50 converted human demonstrations per task achieve a 57\% average success rate, comparable to 52\% with the same number of real-robot demonstrations. Further experiments on spatial and texture editing highlight the potential of \themethod for generating diverse robot demonstrations.
\end{abstract}

%% file: sec/1_intro.tex
\section{Introduction}
\label{sec:intro}

Recent advances in general-purpose robot learning have been driven by Vision-Language-Action~(VLA) models~\cite{intelligence2025pi_, bjorck2025gr00t, kim2024openvla}, whose continued scaling relies on large-scale and diverse real-world demonstrations.
Teleoperated robot demonstrations~\cite{bu2025agibot, khazatsky2024droid, wu2024robomind} preserve real robot-environment interactions and precise low-level control signals, but remain costly and difficult to scale due to hardware and teleoperation requirements.
Egocentric human data~\cite{hoque2026egodex, punamiya2026egoverse, zheng2026egoscale, gao2026dreamdojo, kareer2025emergence} offers a promising alternative.
Such data can be collected from everyday human activities using low-cost VR~\cite{yu2026egokit} and wearable sensing~\cite{wang2026humanego} devices, covering diverse scenes and tasks.
Converting these human experiences into robot-compatible demonstrations offers a direct way to leverage human data for training existing VLA policies without modifying observation and action interfaces.

Existing approaches to human-to-robot demonstration conversion follow two directions.
One line of research~\cite{freeman2026warped, lee2026dex} reconstructs human demonstrations into explicit scene representations to synthesize robot observations through rendering or support policy learning in simulation.
However, these methods may suffer from visual sim-to-real gaps in rendered observations, while pipelines that rely on simulation may additionally face discrepancies in physical dynamics, limiting the fidelity of synthesized robot-environment interactions. 
Another line of research~\cite{lepert2025phantom, wang2026ego2robot} employs video inpainting or generation techniques to synthesize robot demonstrations from egocentric human videos.
Existing video-based approaches generate robot observations from the original egocentric viewpoint, whereas VLA policies~\cite{intelligence2025pi_, bjorck2025gr00t} often rely on simultaneous head-mounted and wrist-mounted observations for precise manipulation.
This observation mismatch limits the direct use of converted demonstrations for training multi-view VLA policies. This raises a fundamental question:
\textit{How can we convert egocentric human demonstrations into multi-view robot demonstrations?}

To address this challenge, we propose \themethod (Figure~\ref{fig:teaser}), a framework that converts egocentric human data into multi-view robot demonstrations for VLA training.
Our key insight is to reverse the direction of data synthesis: starting from existing robot demonstrations, we synthesize corresponding human conditions to supervise human-to-robot multi-view generation without collecting paired data.
Building upon this insight, \themethod consists of three components:
(1) \textbf{Multi-view Video Generation.}
We develop a conditional video generation model that jointly synthesizes head and wrist observations from shared scene geometry and embodiment cues.
Shared geometry provides a common spatial reference across views, while robot renderings and human hand poses guide arm appearance and gripper configurations.
(2) \textbf{Human-Robot Paired Data Construction.}
We construct paired human-robot training data from real multi-view robot demonstrations by mapping robot gripper states to synthetic human hand poses and reconstructing the 4D scenes. 
In this way, we establish cross-embodiment supervision for learning human-to-robot multi-view conversion.
(3) \textbf{End-to-End Demonstration Synthesis.}
To enable practical VLA training, we integrate the learned video generator with robot action retargeting and video-guided gripper-state estimation into an end-to-end demonstration synthesis system.
This system transforms VR-collected human demonstrations into multi-view robot observation-action sequences compatible with existing VLA policies.

We evaluate \themethod on five real-world manipulation tasks spanning rigid, articulated and deformable objects, bimanual manipulation, and multi-stage interactions.
Policies trained on multi-view demonstrations generated from 50 egocentric human demonstrations per task achieve an average success rate of 57\%, comparable to the 52\% obtained using the same number of real-robot demonstrations.
Furthermore, \themethod supports controllable spatial and texture editing, enabling the generation of diverse robot demonstrations.
These results demonstrate the potential of \themethod as a versatile framework for converting human demonstrations into robot-compatible multi-view training data.

%% file: sec/2_related_work.tex
\begin{figure*}[t]
    \centering
    \includegraphics[width=\textwidth]{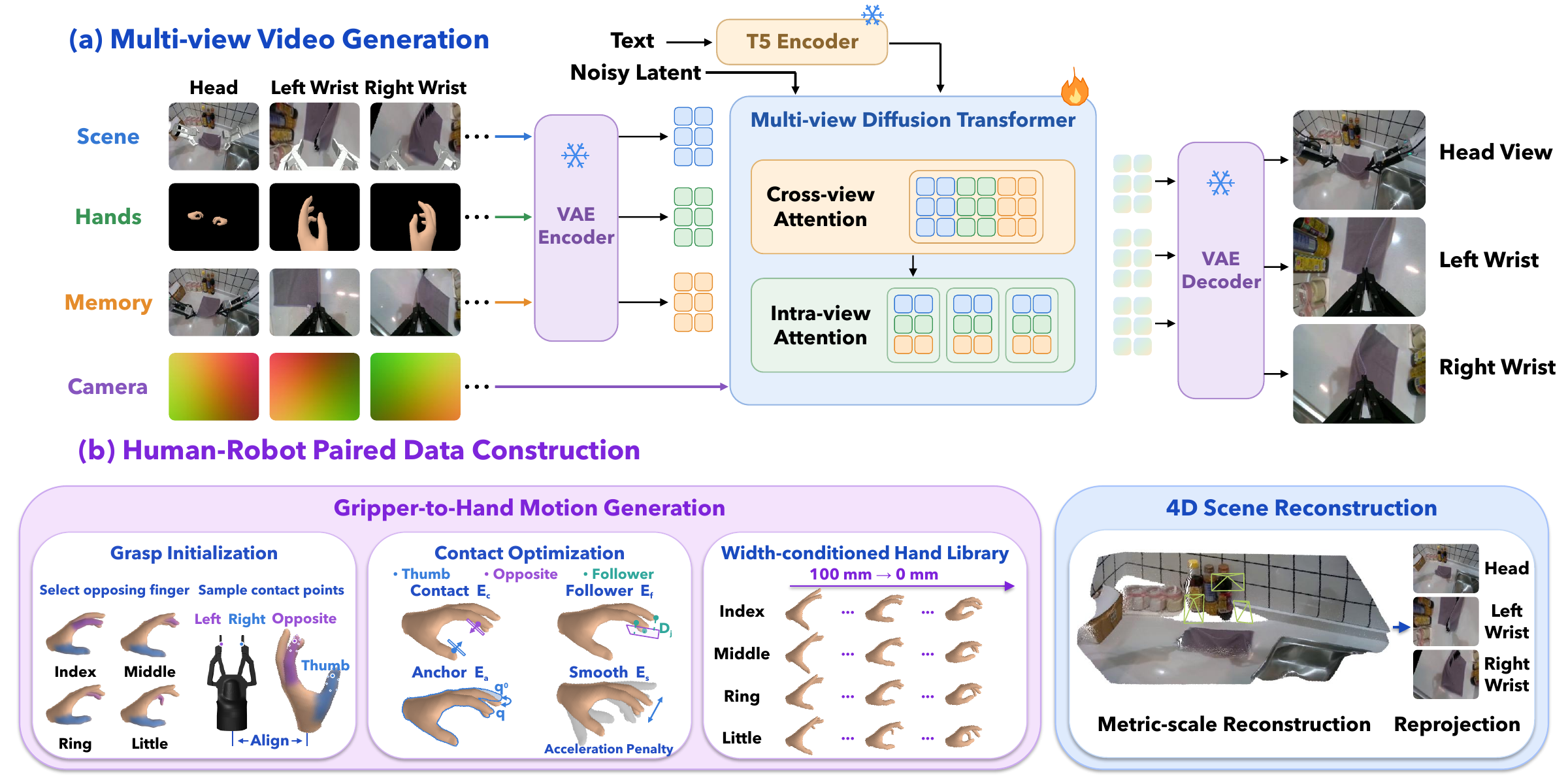}
    \caption{Multi-view video generation and paired data construction in \themethod. (a) Video generation conditioned on 4D scene geometry and hybrid embodiment renderings. (b) Paired data construction through gripper-to-hand motion generation and 4D scene reconstruction.}
        \vspace{-10pt}
    \label{fig:method}
\end{figure*}

\section{Related Work}
\label{sec:related_work}

\subsection{Robot Learning from Egocentric Data}
Egocentric human data~\cite{zheng2026egoscale, punamiya2026egoverse, gao2026dreamdojo, hoque2026egodex, kareer2025emergence, wang2026humanego} has emerged as a promising complement to teleoperated robot demonstrations for scaling robot learning.
Existing approaches to leveraging human data can be broadly categorized into two directions.
One direction~\cite{yuan2026qwen, luo2026being, luo2026beingh07, lepert2026masquerade, zheng2026egoscale, kareer2025egomimic, kareer2025emergence, yuan2026motiontrans, gao2026dreamdojo, ye2026world, ye2025latent, bi2026motus, zhou2026zero} focuses on learning transferable representations or jointly training policies with human and robot data.
These methods leverage large-scale human experiences to improve robotic representations and cross-embodiment knowledge transfer.
In contrast, our work focuses on a data-centric approach that converts human data into robot-compatible demonstrations, allowing them to be directly incorporated into existing VLA training. 

Another direction~\cite{lepert2025phantom, wang2026ego2robot, freeman2026warped, yu2025real2render2real, li2025mimicdreamer, heng2025rwor, liu2026egoengine, strong2026real2sem2real, wang2026dexagent, lee2026dex, ci2025h2r, li2025h2r} directly converts egocentric human demonstrations into robot training data.
Scene rendering and simulation methods~\cite{freeman2026warped, yu2025real2render2real, liu2026egoengine, strong2026real2sem2real, wang2026dexagent, lee2026dex} recover human-object interactions and scene geometry to synthesize robot demonstrations through rendering or support policy learning in simulation.
Although recent Real2Sim works~\cite{chen2026agentic, chen2026robosimstudio, ranawaka2026simfoundry, pfaff2026scenesmith, jiang2026gsworld, jain2025polaris, fan2025twinaligner} have improved reconstruction and simulation fidelity, rendered observations may still exhibit visual sim-to-real gaps, while simulation-based pipelines additionally face discrepancies in physical dynamics.
Video editing and generation methods~\cite{lepert2025phantom, wang2026ego2robot, li2025mimicdreamer, ci2025h2r, lepert2026masquerade} translate egocentric human videos into robot observations through inpainting, robot compositing, or learned video generation.
However, existing approaches primarily focus on synthesizing observations within the original egocentric viewpoint, leaving a mismatch with the multi-view observation requirements of modern VLA policies~\cite{intelligence2025pi_, bjorck2025gr00t}.
In contrast, \themethod learns to generate multi-view robot observations through paired human-robot supervision constructed in the reverse direction, and integrates the generated observations with retargeted robot actions and video-guided gripper-state estimation to produce complete VLA-compatible demonstrations.

\subsection{Controllable Video Generation}

We condition multi-view video generation on scene geometry, camera poses, and embodiment cues to control scene content, viewpoints, and robot motion.
Camera-controlled video generation~\cite{lin2026vista4d, cao2026freeorbit4d, yang2026neoverse,jiang2025vace,van2026anyview,liu2025worldmirror,xie2026lavr,yu2025trajectorycrafter,ren2025gen3c,bai2025recammaster,qian2025wristworld,zeng2026ego2wrist} has advanced rapidly with the development of video foundation models~\cite{wan2025wan,yang2025cogvideox,seedance2026seedance}.
Implicit methods~\cite{jiang2025vace,van2026anyview,xie2026lavr,bai2025recammaster} learn multi-view consistency directly from data, but scaling real-world video datasets with precise camera poses remains challenging, potentially limiting their generalization ability.
Explicit methods~\cite{lin2026vista4d,cao2026freeorbit4d,yang2026neoverse,liu2025worldmirror,yu2025trajectorycrafter,ren2025gen3c,qian2025wristworld} lift videos into 4D point clouds, reproject them into target viewpoints, and condition novel-view video generation on the resulting projections.
Feed-forward reconstruction models~\cite{wang2025vggt,wang2026vggt,li2025megasam,zhang2026efficiently,lin2025depth} support the scalability of geometry-guided approaches.
Inspired by explicit methods, \themethod reconstructs 4D scene geometry from egocentric human demonstrations and reprojects it into robot head and wrist views. These projections, together with embodiment cues, condition joint multi-view robot video generation.

Meanwhile, robot-controlled video generation~\cite{alzayer2026masked,huang2026roborender,gu2026geniworld,wu2026oscar,ye2026anchordream,zhao2026real2edit2real,jiang2025enerverse,zhao2026rynnworld,guo2026ctrl} has also advanced with the development of robotic world foundation models~\cite{agarwal2026cosmos,ye2026world,qiu2026ge,liao2026genie}.
Compared with methods that use implicit robot-state representations~\cite{jiang2025enerverse,zhao2026rynnworld,guo2026ctrl} as control signals, explicit approaches~\cite{alzayer2026masked,huang2026roborender,gu2026geniworld,wu2026oscar,ye2026anchordream,zhao2026real2edit2real} render robot arms as visual conditions, providing explicit visual guidance for robot appearance and motion.
To generate robot videos directly from egocentric human data, \themethod extends these visual conditions to incorporate both robot arms and human hands.
This design enables the model to learn the correspondence between human hand poses and robot grippers while preserving the quality of robot arm generation.

%% file: sec/3_method.tex
\section{Method}
\label{sec:method}

\subsection{Overview}
\label{subsec:overview}

We consider the task of converting egocentric human data into multi-view robot demonstrations for downstream VLA training.
Given a stereo egocentric human video
$\mathbf{I}_H = \{(I_{H,l}^t, I_{H,r}^t)\}_{t=1}^{T}$
and a target robot configuration $\mathcal{K}$,
we formulate the conversion as:
\begin{equation}
    \mathbf{O}_R =
    \mathcal{F}_{\theta}(\mathbf{I}_H, \mathcal{K})
    = (\mathbf{I}_R, \mathbf{q}_R, \mathbf{a}_R),
    \label{eq:problem}
\end{equation}
where $\mathbf{I}_R = \{(I_h^t, I_l^t, I_r^t)\}_{t=1}^{T'}$
denotes the generated head, left-wrist, and right-wrist robot videos,
while $\mathbf{q}_R$ and $\mathbf{a}_R$ denote the corresponding joint states and actions.
The robot configuration $\mathcal{K} = (\mathcal{K}_{robot}, \mathcal{K}_{cam})$ specifies the robot URDF model and camera parameters.

\begin{figure*}[t]
    \centering
    \includegraphics[width=\textwidth]{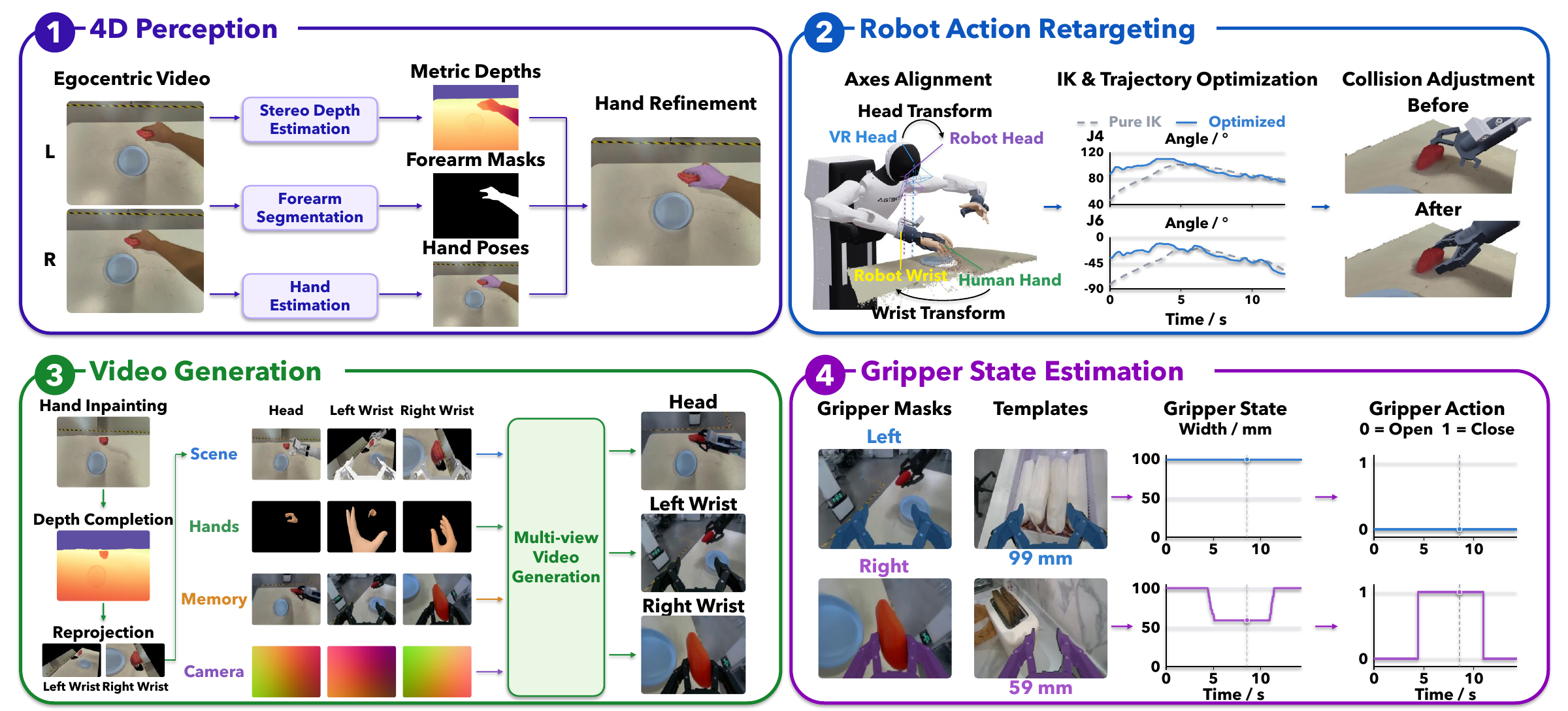}
    \caption{The framework of end-to-end demonstration synthesis. (1) 4D perception from binocular egocentric videos. (2) Retargeting robot actions from human hand poses. (3) Generating multi-view robot videos. (4) Estimating gripper states from generated videos.}
        \vspace{-10pt}
    \label{fig:inference}
\end{figure*}

As shown in Figures~\ref{fig:method} and~\ref{fig:inference}, \themethod comprises three components: multi-view video generation, human-robot paired data construction, and end-to-end demonstration synthesis.
In Sec.~\ref{subsec:model}, we introduce a multi-view video generator $\mathcal{G}_{\theta}$ conditioned on robot configuration $\mathcal{K}$, joint states $\mathbf{q}_R$, 4D scene $\mathbf{s}$, and human hand poses $\mathbf{h}$:
\begin{equation}
    \mathbf{I}_R =
    \mathcal{G}_{\theta}
    (\mathcal{K}, \mathbf{q}_R, \mathbf{s}, \mathbf{h}).
    \label{eq:gen_def}
\end{equation}
Since existing robot demonstration datasets~\cite{bu2025agibot,khazatsky2024droid} lack the required scene and hand conditions, Sec.~\ref{subsec:data} presents a reverse data construction pipeline $\mathcal{P}$ that constructs these conditions from real multi-view robot demonstrations $\mathbf{O}_R^{real}$:
\begin{equation}
    (\mathbf{s}, \mathbf{h}) =
    \mathcal{P}(\mathcal{K}, \mathbf{O}_R^{real}).
    \label{eq:paired_def}
\end{equation}
We pair these scene and hand-pose conditions with the original multi-view robot videos to train $\mathcal{G}_{\theta}$.
Finally, Sec.~\ref{subsec:inference} details the end-to-end demonstration synthesis system including 4D perception, robot action retargeting, multi-view video generation, and gripper-state estimation to realize $\mathcal{F}_{\theta}$.

\subsection{Multi-view Video Generation}
\label{subsec:model}

As shown in Figure~\ref{fig:method}(a), we develop a multi-view diffusion transformer based on~\cite{liao2026genie,ali2025world}, leveraging a Cosmos-Predict backbone pretrained on robot demonstrations~\cite{bu2025agibot}.

\noindent\textbf{Visual Conditions.} We extend robot-controlled video generation with camera-controlled scene geometry and hybrid human-robot embodiment conditioning.
A 4D point cloud represents the background and manipulated objects while excluding robot arms and human hands, decoupling scene geometry from embodiment. For each viewpoint $v\in\{h,l,r\}$, we construct two visual conditions:
\begin{equation}
\begin{aligned}
    \mathbf{C}_{scene}^{v}
    &= \Sigma\big(
    \Pi_v(\mathbf{s}),
    \mathcal{R}_{\text{URDF}}(\mathcal{K}^{v},\mathbf{q}_R)
    \big),\\
    \mathbf{C}_{hand}^{v}
    &= \mathcal{R}_{\text{MANO}}(\mathcal{K}^{v},\mathbf{h}).
\end{aligned}
\label{eq:visual_conditions}
\end{equation}
Here, $\Pi_v$ projects the shared 4D scene into viewpoint $v$, $\mathcal{R}_{\text{URDF}}$ renders the robot with an open gripper, and $\Sigma$ denotes depth-aware compositing.
$\mathcal{R}_{\text{MANO}}$ renders the posed human hands~\cite{MANO:SIGGRAPHASIA:2017} under the corresponding camera parameters.
Shared geometry provides a common spatial reference across views and supports spatial and texture augmentation through point-cloud edits~(Sec.~\ref{subsec:aug}).
Meanwhile, the hybrid embodiment conditions decouple robot arm and gripper control: URDF rendering specifies the robot arm and gripper base, while human hand poses guide the gripper's visual configuration.
The model learns to infer gripper visual configurations from hand-pose conditions, without a manually specified finger-to-gripper-width rule at inference.

\noindent\textbf{Model Architecture.}
We adapt the pretrained video prediction backbone through full-chunk conditioning, providing visual guidance at every frame rather than only the first frame.
The scene, hand, and historical RGB conditions are encoded by the pretrained VAE encoder~\cite{wan2025wan}, while camera ray origin and direction maps provide explicit geometric conditioning without VAE encoding.
The multi-view diffusion transformer alternates cross-view and intra-view attention to exchange information across viewpoints while modeling view-specific details.
For long-horizon synthesis, we retain the backbone's autoregressive memory, which conditions each chunk on previously generated frames.

\noindent\textbf{Memory Dropout.}
Training with ground-truth memory frames may cause the model to rely on memory shortcuts rather than visual conditions.
This is particularly problematic during autoregressive inference, where the first chunk lacks robot history and subsequent chunks may accumulate generation errors.
We therefore randomly replace the memory with Gaussian noise during training:
\begin{equation}
    \mathbf{M}_{train} =
    \begin{cases}
        \boldsymbol{\epsilon}, & u < p,\\
        \mathbf{M}_{GT}, & u \geq p,
    \end{cases}
    \label{eq:memory_dropout}
\end{equation}
where $u\sim\mathcal{U}(0,1)$, $\boldsymbol{\epsilon}\sim\mathcal{N}(\mathbf{0},\mathbf{I})$, $\mathbf{M}_{GT}$ is the ground-truth memory and $p\in[0,1]$ denotes the dropout probability.
This encourages the model to use visual conditions when valid memory is unavailable, improving first-chunk generation.

\subsection{Human-Robot Paired Data Construction}
\label{subsec:data}

As illustrated in Figure~\ref{fig:method}(b), we construct paired human-robot demonstrations through reverse synthesis, reconstructing 4D scenes and synthesizing corresponding human hand motions from robot demonstrations.

\noindent\textbf{Gripper-to-Hand Motion Generation.}
To diversify human hand conditions during training, we construct a library of 2,000 grasp styles, each defining a hand-pose trajectory parameterized by gripper width.
A fixed rigid transformation aligns the robot flange with the human wrist.
In the flange frame, the thumb corresponds to one gripper side, while the opposing finger is sampled from the index, middle, ring, and little fingers.
Each grasp style is initialized by sampling contact points on the finger surfaces and randomizing joint angles.
The hand trajectories are optimized to maintain contact and smoothness across gripper widths.

Let $w_k$ denote the gripper width, $\mathbf{q}_k$ the hand joint angles, and $\mathbf{q}_{\mathcal{W}}=\{\mathbf{q}_k\}_{k\in\mathcal{W}}$ the poses within a local width window $\mathcal{W}$.
We optimize:
\begin{equation}
\begin{aligned}
\min_{\mathbf{q}_{\mathcal{W}}}\quad
&\sum_{k\in\mathcal{W}}
(\lambda_c E_c^k+\lambda_f E_f^k
+\lambda_a E_a^k)
+\lambda_s E_s\\
\text{s.t.}\quad
&\mathbf{q}_{\min}\preceq\mathbf{q}_k
\preceq\mathbf{q}_{\max},\\
&|\mathbf{q}_{k+1}-\mathbf{q}_k|
\preceq\boldsymbol{\delta}_k,
\end{aligned}
\label{eq:hand_synthesis}
\end{equation}
where $\mathbf{q}_{\min}$ and $\mathbf{q}_{\max}$ specify anatomical joint limits, and $\boldsymbol{\delta}_k$ bounds joint-angle changes between adjacent widths.

The contact loss $E_c^k$ aligns the selected finger contacts with the gripper surfaces:
\begin{equation}
\begin{aligned}
E_c^k=\sum_{i}\big[
&\operatorname{dist}^2(
\mathbf{p}_i(\mathbf{q}_k),\mathcal{F}_i(w_k))\\
&+\lambda_n
\max(0,\eta-\mathbf{n}_i(\mathbf{q}_k)^\top
\bar{\mathbf{n}}_i)^2\big],
\end{aligned}
\end{equation}
where $i\in\{\text{thumb}, \text{opposite}\}$ indexes the thumb and opposing finger, with contact positions $\mathbf{p}_i$ and normals $\mathbf{n}_i$.
$\mathcal{F}_i(w_k)$ denotes the corresponding gripper-finger contact region at width $w_k$, $\bar{\mathbf{n}}_i$ its opposing surface normal, and $\eta$ the minimum normal-alignment threshold.

The follower loss $E_f^k$ encourages the remaining three fingers to maintain fixed distances from their corresponding gripper-side planes throughout closing, while the anchor loss $E_a^k$ discourages deviations from the initial hand pose:
\begin{equation}
\begin{aligned}
E_f^k&=\sum_{j\in\mathcal{U}}
\operatorname{SmoothL1}(D_j^k-\bar{d}_j),\\
E_a^k&=\|\mathbf{q}_k-\mathbf{q}_k^0\|_2^2,
\end{aligned}
\end{equation}
where $\mathcal{U}$ denotes the three non-contact fingers, $D_j^k$ is the signed distance from a selected finger point to its corresponding gripper-side plane, and $\bar{d}_j$ is its fixed target distance.
$\mathbf{q}_k^0$ denotes the initial hand pose.

To ensure smooth closing motions, we progressively decrease gripper width from $100$ to $0\,\mathrm{mm}$, initializing each local window from preceding solutions. The smoothness loss $E_s$ is:
\begin{equation}
E_s=\sum_k\|\mathbf{v}_{k+1}-\mathbf{v}_k\|_2^2,
\end{equation}
where $\mathbf{v}_k=(\mathbf{q}_{k+1}-\mathbf{q}_k)/(w_{k+1}-w_k)$.
We optimize over four-sample windows with adaptive step refinement and sample the resulting trajectories at $2\,\mathrm{mm}$ intervals.
During training, a grasp style is randomly selected for each iteration, and its hand poses are interpolated according to the recorded gripper widths.

\noindent\textbf{4D Scene Reconstruction.}
To recover the 4D scene, we first employ a metric-scale VGGT model~\cite{zhao2026real2edit2real} to jointly reconstruct the head, left wrist, and right wrist observations.
We then render the robot URDF model into these viewpoints to obtain 2D bounding boxes of the robot arms, which guide robot segmentation using SAM-2~\cite{ravi2025sam}.
Finally, we remove the robot regions from the reconstructed RGB-D observations and use Minimax-Remover~\cite{zi2026minimax} and Lingbot-Depth~\cite{tan2026masked} to inpaint the occluded RGB regions and complete the corresponding depth maps, respectively.
The recovered RGB-D observations provide the scene geometry required for subsequent viewpoint reprojection.

\input{tables/maintable}

\subsection{End-to-end Demonstration Synthesis}
\label{subsec:inference}

Given the trained multi-view video generation model, we develop an end-to-end system for synthesizing robot demonstrations from VR-collected human data, as illustrated in Figure~\ref{fig:inference}.
Human demonstrations are captured using PICO 4 Ultra~\cite{pico2026pico4ultra} and its spatial computing SDK~\cite{pico2026spatialmp4}, providing stereo RGB videos with calibrated camera poses.

\noindent\textbf{4D Perception.}
Stereo depths are estimated using Fast-FoundationStereo~\cite{wen2026fast}, followed by depth completion with Lingbot-Depth.
Meanwhile, SAM-3~\cite{carion2026sam} provides forearm segmentation, and HaWoR~\cite{zhang2025hawor} estimates hand poses.
The estimated hand poses are further refined against the recovered RGB-D geometry to improve spatial alignment.   

\noindent\textbf{Robot Action Retargeting.}
We first align the human and robot head-camera coordinate frames by matching their estimated vertical distances to the tabletop.
A fixed transformation between human wrist poses and robot end-effector poses is then established to transfer human motions into the robot workspace.
Based on these mappings, inverse kinematics is solved with cuRobo~\cite{sundaralingam2023curobo}, followed by joint trajectory optimization with Clarabel~\cite{goulart2026clarabel} to improve motion accuracy and temporal smoothness.
Potential collisions with the estimated tabletop are further resolved by iteratively adjusting the flange orientation and target end-effector height.

\noindent\textbf{Video Generation.}
To construct 4D scene conditions, the human forearms are first removed from the head-view observations using Minimax-Remover, followed by depth completion with Lingbot-Depth.
The recovered 4D scene is then reprojected into the target viewpoints defined by the URDF camera configurations.
Given the retargeted robot trajectories, URDF robot arms are rendered across all viewpoints and combined with the projected scene.
Together with rendered hand poses and camera conditions, these visual conditions are fed into the multi-view video generation model to synthesize the robot observations.

\noindent\textbf{Gripper State Estimation.}
To recover gripper states from generated robot videos, we use SAM-2 with fixed point prompts to segment grippers in the wrist views.
The same procedure is applied to robot dataset observations with known gripper states to establish a mask-to-state mapping.
We then use this mapping to estimate gripper states from the generated wrist-view videos, completing the multi-view robot observation-action sequences for VLA training.

%% file: tables/maintable.tex
\begin{table*}[t]
\renewcommand\tabcolsep{6pt}
\centering
\begin{tabular}{l|ccccc|c}
\toprule
\textbf{Method}
& \textbf{Pick-Place}
& \textbf{Pour Water}
& \textbf{Pack Vegetables}
& \textbf{Fold Towel}
& \textbf{Fold Clothes}
& \textbf{Total} \\
\midrule
Phantom~\cite{lepert2025phantom}
& 0 / 20 & 2 / 20 & 0 / 20 & 1 / 20 & 0 / 20 & 3\% \\
Ego2Robot~\cite{wang2026ego2robot}
& 5 / 20 & 5 / 20 & 3 / 20 & 2 / 20 & 0 / 20 & 15\% \\
\midrule
Real (Single-view)
& 4 / 20 & 4 / 20 & 8 / 20 & 4 / 20 & 6 / 20 & 26\% \\
Ours (Single-view)
& 11 / 20 & 7 / 20 & 3 / 20 & 4 / 20 & 9 / 20 & 34\% \\
\midrule
Real (Multi-view)
& 12 / 20 & \textbf{10 / 20} & \textbf{11 / 20}
& 9 / 20 & \textbf{10 / 20} & 52\% \\
Ours (Multi-view)
& \textbf{14 / 20} & \textbf{10 / 20} & 10 / 20
& \textbf{13 / 20} & \textbf{10 / 20} & \textbf{57\%} \\
\bottomrule
\end{tabular}
\caption{
Success rates of \themethod on five real-world robotic manipulation tasks. The policy trained on data generated from 50 human demonstrations per task can achieve performance comparable to that of a policy trained on the same number of real-robot demonstrations.
}
        \vspace{-10pt}
\label{tab:real_robot_results}
\end{table*}

%% file: sec/4_experiment.tex
\section{Experiments}
\label{sec:experiment}

\subsection{Implementation Details}

\noindent\textbf{\themethod.}
For multi-view video generation, we uniformly sample 15K episodes spanning 147 tasks and 224 hours from Agibot-World~\cite{bu2025agibot}, with a training-to-validation split of 9:1. The video model's training data have no scene overlap with our real-world evaluation environments.
We adapt the GE-Sim~\cite{liao2026genie} backbone and fully fine-tune the diffusion transformer for 12K iterations using 480 A100 GPU hours, a global batch size of 16, and a learning rate of $10^{-4}$.
The memory dropout probability $p$ is set to 0.5.
We use 35 denoising steps for training-time sampling and 10 for inference.
Generating a 20-second three-view video at 30 fps typically takes 82.8 seconds on 8 A100 GPUs.
Further details are provided in the supplementary material.

\noindent\textbf{Hardware.}
We deploy our policies on the Agibot Genie G1 robot, equipped with a RealSense D455 head camera and two D405 wrist cameras.
The workspace covers a $50\,\mathrm{cm}\times30\,\mathrm{cm}$ rectangular tabletop region.
Additional details are provided in the supplementary material.

\noindent\textbf{VLA Policy.}
We evaluate the quality of our generated demonstrations using the VLA policy $\pi_{0.5}$~\cite{intelligence2025pi_}.
Training uses a global batch size of 256 and typically requires 10K iterations and approximately 100 A100 GPU hours.

\subsection{Real-World Policy Evaluation}

\noindent\textbf{Tasks and Settings.} As shown in Figure~\ref{fig:teaser}, we evaluate five real-world manipulation tasks involving rigid, articulated, and deformable objects, as well as bimanual and multi-stage interactions.
For each task, we collect 50 egocentric human demonstrations and compare our policies with those trained on 50 real-robot demonstrations and data converted by Phantom~\cite{lepert2025phantom} and Ego2Robot~\cite{wang2026ego2robot}.
Each policy is evaluated over 20 trials, with results reported in Table~\ref{tab:real_robot_results}.

\begin{itemize}
    \item \textbf{Pick-Place:} The robot grasps the red sweet potato with its right arm and places it on the blue plate.
    \item \textbf{Pour Water:} The robot grasps the yellow cup with its left arm, aligns it with the green cup, and pours water.
    \item \textbf{Pack Vegetables:} The robot's left arm puts the green vegetable in the basket and closes the lid without snagging.
    \item \textbf{Fold Towel:} The robot folds the towel with both arms.
    \item \textbf{Fold Clothes:} The robot folds the clothes with both arms.
\end{itemize}

\begin{figure*}[t]
    \centering
    \includegraphics[width=\textwidth]{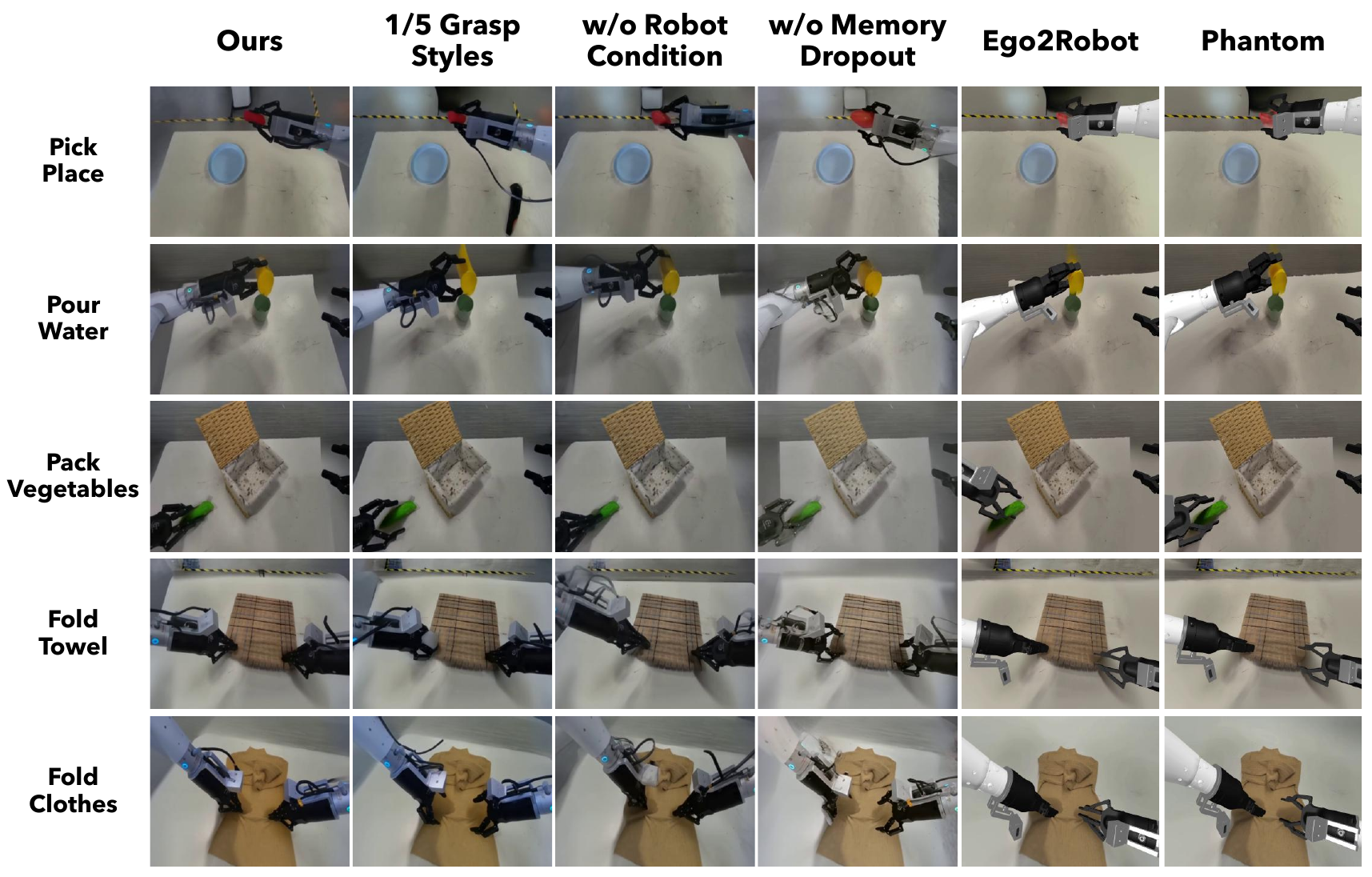}
    \caption{Qualitative baseline comparisons and ablation study.}
    \vspace{-5pt}
    \label{fig:ablation}
\end{figure*}

\noindent\textbf{Comparison with Real-world Data.}
Across the five tasks, our method achieves an average success rate of 57\%, comparable to the 52\% obtained with real-robot demonstrations.
These results demonstrate that multi-view robot demonstrations synthesized from egocentric human data can provide effective supervision for downstream VLA training.

\noindent\textbf{Comparison with Baselines.}
Phantom and Ego2Robot achieve average success rates of 3\% and 15\%, respectively, compared with 34\% for our single-view variant and 57\% for our multi-view system.
Notably, our single-view variant outperforms both baselines, suggesting that the gains are not solely attributable to wrist observations.
Our framework retargets human wrist poses to robot flange poses and estimates gripper states from generated videos, without explicitly computing gripper width from finger distances at inference.
Figure~\ref{fig:ablation} compares converted observations.

\input{tables/ablation}

\noindent\textbf{Impact of Multi-view Observations.}
Multi-view training consistently improves policy performance for both real and generated demonstrations, increasing average success rates from 26\% to 52\% and from 34\% to 57\%, respectively.
The largest gains are observed on Pack Vegetables and Fold Towel, with increases of 35 and 45 percentage points, respectively. These gains may reflect the importance of near observations for contact-rich and bimanual manipulation.

\subsection{Ablation Study}

\begin{figure*}[htbp]
    \centering
    \includegraphics[width=\textwidth]{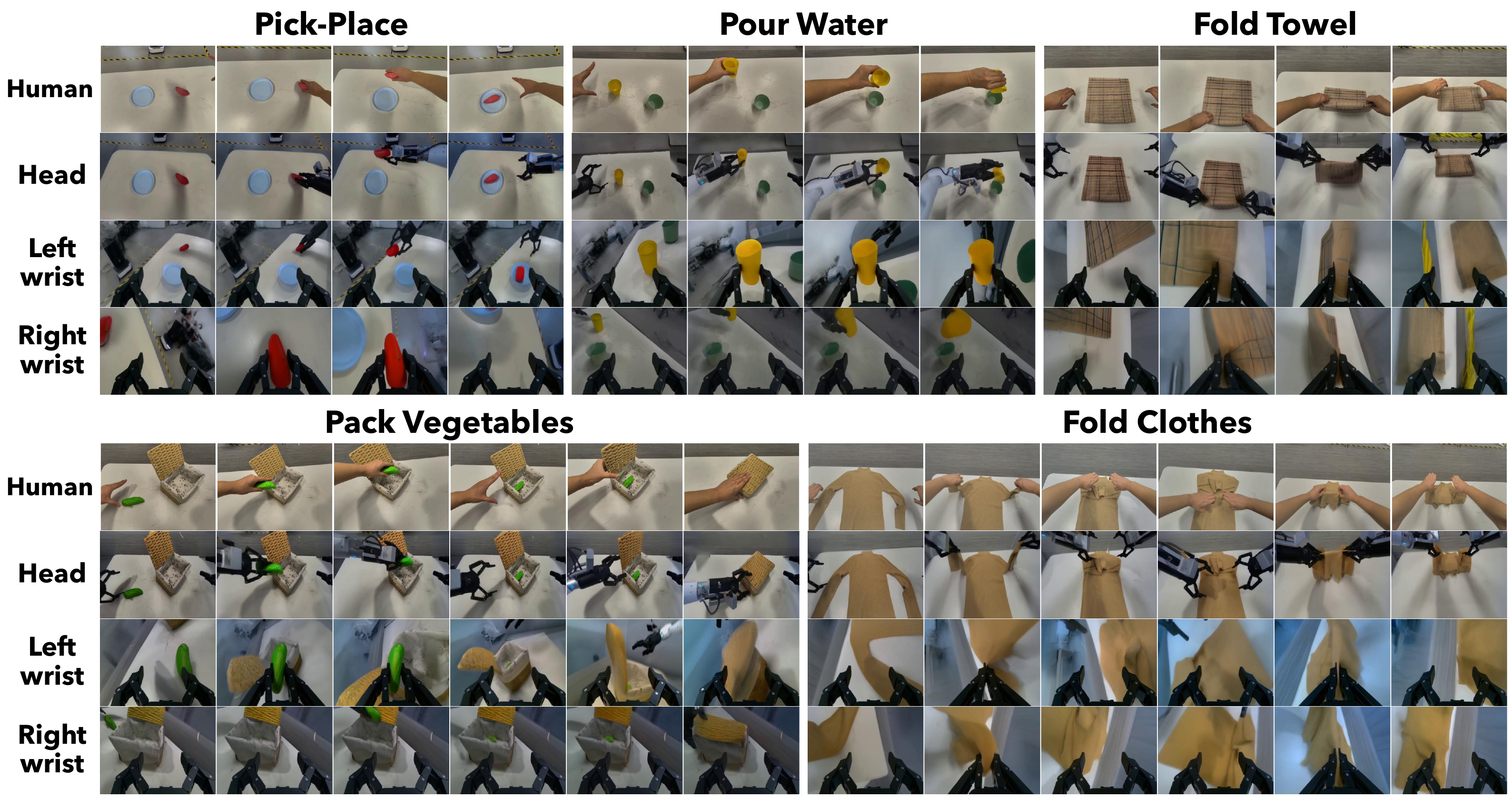}
    \caption{Visualization of videos generated by \themethod on all five real-world tasks.}
    \label{fig:vis}
\end{figure*}

\noindent\textbf{Settings.}
We randomly sample 4,000 validation video chunks to evaluate generation quality without memory and qualitatively assess generation on egocentric data.

\noindent\textbf{Effect of Grasp Styles.}
We reduce the grasp library to one-fifth of its original size.
The ablation variant is validated using grasp styles from the reduced training library, yielding similar in-distribution video quality to the full method in Table~\ref{tab:ablation_metrics}.
On egocentric human demonstrations, the reduced-library variant produces incorrect gripper widths in the examples shown in Figure~\ref{fig:ablation}, suggesting that greater grasp diversity may improve transfer to real human hand poses.

\noindent\textbf{Effect of Robot Condition.}
We remove the URDF robot renderings from the scene conditions to examine their role in guiding robot appearance.
As shown in Figure~\ref{fig:ablation}, the model still generates plausible gripper configurations from the hand-pose conditions, but the robot appearance becomes inconsistent.
The degraded video metrics in Table~\ref{tab:ablation_metrics} further support the importance of explicit robot conditioning.

\noindent\textbf{Effect of Memory Dropout.}
We train a variant without memory dropout by setting $p=0$.
When historical memory is unavailable, this variant exhibits a substantial degradation in generation quality (Table~\ref{tab:ablation_metrics}).
The qualitative results in Figure~\ref{fig:ablation} also show visual artifacts and incorrect gripper widths.
These results suggest that training with historical frames alone encourages excessive reliance on memory, whereas memory dropout helps the model use the scene and hand-pose conditions when generating the first chunk.

\subsection{Controllable Data Augmentation}
\label{subsec:aug}

\noindent\textbf{Spatial Editing.}
We use SAM-3~\cite{carion2026sam} to segment target objects, translate their point clouds by random distances of $2$--$5\,\mathrm{cm}$, and apply the same transformations to the hand poses to maintain spatial alignment.
Our synthesis then processes the edited data.
Table~\ref{tab:data_scaling} shows that augmenting 50 demonstrations to 150 raises average success on Pick-Place and Pour Water from 60\% to 82.5\%, demonstrating the benefit of spatial augmentation.
See Figure~\ref{fig:teaser} for examples.

\input{tables/spatialaug}

\noindent\textbf{Texture Editing.}
We use JoyAI-Video-Edit~\cite{xiao2026joyai} to edit egocentric video textures while preserving robot trajectories and human hand poses.
The edited scene conditions enable multi-view demonstration generation with new appearances.
As shown in Table~\ref{tab:texture_generalization}, 
texture-adapted data synthesis improves the average success rate on Fold Clothes from 35\% to 50\% across five tabletop textures, showing the effectiveness of editing. See Figure~\ref{fig:teaser} for examples.

\subsection{Visualization}

Figure~\ref{fig:vis} visualizes generated videos across five manipulation tasks, illustrating visual fidelity, human-to-robot motion alignment, and multi-view consistency.
Additional results are provided in the supplementary material.

\input{tables/texaug}

%% file: tables/ablation.tex
\begin{table}[t]
\renewcommand\tabcolsep{2pt}
\centering
\begin{tabular}{lcccc}
\toprule
\textbf{Model}
& \textbf{PSNR} $\uparrow$
& \textbf{SSIM} $\uparrow$
& \textbf{LPIPS} $\downarrow$
& \textbf{FVD} $\downarrow$ \\
\midrule
Ours
& \textbf{16.858} & \textbf{0.6606}
& \textbf{0.3153} & 66.15 \\
1/5 Grasp Styles
& 16.609 & 0.6515 & 0.3205 & \textbf{60.99} \\
w/o Robot Condition
& 15.661 & 0.6202 & 0.3626 & 73.38 \\
w/o Memory Dropout
& 13.521 & 0.5964 & 0.5169 & 1117.30 \\
\bottomrule
\end{tabular}
\caption{
Quantitative ablation study on 4,000 chunks sampled from the validation set.
}
\label{tab:ablation_metrics}
\vspace{-10pt}
\end{table}

%% file: tables/spatialaug.tex
\begin{table}[t]
\renewcommand\tabcolsep{5pt}
\centering
\begin{tabular}{lccc}
\toprule
\textbf{\# Demo} & \textbf{Pick-Place} & \textbf{Pour Water} & \textbf{Total} \\
\midrule
Real 50
& 12 / 20 & 10 / 20 & 55.0\% \\
Ours 50
& 14 / 20 & 10 / 20 & 60.0\% \\
Ours 50 Aug 100
& 15 / 20 & 12 / 20 & 67.5\% \\
Ours 50 Aug 150
& 17 / 20 & 16 / 20 & 82.5\% \\
\bottomrule
\end{tabular}
\caption{
Performance comparison with different amounts of spatially augmented training data on the Pick-Place and Pour Water tasks. Ours 50 Aug 100/150 denotes a total of 100/150 demonstrations
obtained by augmenting 50 human demonstrations.
}
\label{tab:data_scaling}
\end{table}

%% file: tables/texaug.tex
\begin{table}[t]
\renewcommand\tabcolsep{2pt}
\centering
\begin{tabular}{lcccccc}
\toprule
\textbf{\# Demo}
& \textbf{White}
& \textbf{Blue}
& \textbf{Flower}
& \textbf{Grid}
& \textbf{Green}
& \textbf{Total} \\
\midrule
Real 50
& 10 / 20 & 7 / 20 & 9 / 20 & 7 / 20 & 7 / 20 & 40\% \\
Ours 50
& 10 / 20 & 7 / 20 & 7 / 20 & 5 / 20 & 6 / 20 & 35\% \\
Ours 50$^*$
& 10 / 20 & \textbf{10 / 20} & \textbf{11 / 20}
& \textbf{10 / 20} & \textbf{9 / 20} & \textbf{50\%} \\
\bottomrule
\end{tabular}
\caption{
Performance comparison under different background textures on the Fold Clothes task. Ours 50* denotes that the policy is separately trained on 50 texture-adapted demonstrations.
}
\label{tab:texture_generalization}
\vspace{-15pt}
\end{table}

%% file: sec/5_conclusion.tex
\section{Conclusion}
\label{sec:conclusion}

We present \themethod, a framework for converting egocentric human demonstrations into multi-view robot demonstrations.
Through reverse data synthesis, \themethod learns human-to-robot multi-view generation without collecting paired human-robot data, and combines generated observations with retargeted actions for policy training.
Experiments on five real-world manipulation tasks demonstrate that policies trained on our generated multi-view demonstrations perform comparably to those trained on an equal number of real-robot demonstrations.
Further experiments on spatial and texture editing highlight the potential of \themethod as a versatile robot demonstration generator.

%% file: sec/llm_usage.tex
\section*{LLM Usage Statement}

OpenAI ChatGPT was used to assist with language editing, manuscript organization, and figure design and editing. OpenAI Codex was additionally used for code implementation and debugging. All AI-assisted content and code were reviewed and validated by the authors, who take full responsibility for the paper's technical claims, citations, and experimental results.

%% file: sec/ack.tex
\section*{Acknowledgement}
\label{sec:ack}
\begin{sloppypar}
We would like to thank Hao Huang from Peking University and Zhexuan Yang from ArcheBase for their technical supports.
\end{sloppypar}

%% file: sec/X_suppl.tex
\clearpage

\input{tables/methodcomparison}

\section{Method Comparison}

To further clarify the distinction between \themethod and prior
human-to-robot demonstration conversion methods, we provide
a systematic comparison in Table~\ref{tab:supp_compare}.

\noindent\textbf{Scene Rendering and Simulation Methods.}
Methods such as WARPED~\cite{freeman2026warped},
EgoEngine~\cite{liu2026egoengine},
Real2Render2Real~\cite{yu2025real2render2real}, and
Real2Sem2Real~\cite{strong2026real2sem2real}
leverage 3D assets to synthesize robot observations.
Most of these approaches offer geometric control and support scene editing and, in some cases, wrist-view rendering.
However, preparing 3D assets introduces additional scene-dependent requirements.

\noindent\textbf{Video Editing and Generation Methods.}
H2R~\cite{li2025h2r}, Phantom~\cite{lepert2025phantom},
Masquerade~\cite{lepert2026masquerade},
H2R-Grounder~\cite{ci2025h2r},
MimicDreamer~\cite{li2025mimicdreamer}, and
Ego2Robot~\cite{wang2026ego2robot}
convert human observations into robot-compatible
visual data without 3D assets.
However, these approaches primarily focus on the
original egocentric viewpoint, without demonstrating
synchronized generation of robot head and wrist views. 

\noindent\textbf{\themethod.}
Unlike prior approaches, \themethod jointly generates consistent robot head and dual-wrist observations
without requiring 3D scene assets.
Through reverse synthesis, we construct human-robot
paired supervision from existing robot demonstrations,
avoiding paired human-robot data collection.
Furthermore, \themethod supports controllable scene
editing to augment the generated robot demonstrations with spatial and texture variations.

\section{\themethod Technical Details}

\subsection{Multi-view Video Generation}
\noindent\textbf{Camera-Robot Joint Refinement.}
Errors in recorded camera calibration and joint angles can misalign
robot renders with the observed images
\cite{bu2025agibot,khazatsky2024droid}. Inspired by
PointWorld~\cite{Huang_2026_CVPR}, we refine camera parameters and
joint offsets by aligning rendered robot silhouettes with
observed segmentation masks.

For each camera view, let $\mathbf{q}_t$ denote the
recorded joint angles at frame $t$, $\boldsymbol{\theta}$ the
correction vector shared across frames of the demonstration, and
$a\in\{L,R\}$ the index of the physical robot arm and its gripper.
We use forward kinematics~\cite{sundaralingam2023curobo} and
differentiable rendering~\cite{laine2020nvdiffrast} to obtain the
rendered mask $\widehat{M}_{t,a}$. We obtain the corresponding
observed mask $M_{t,a}$ by prompting SAM-2~\cite{ravi2025sam} with
bounding boxes from the robot render. The head view uses arm masks,
and each wrist view uses the masks of the grippers visible in that
camera.

We minimize
\begin{equation}
\begin{aligned}
\min_{\boldsymbol{\theta}\in\mathcal{B}}\quad
&\frac{1}{|\mathcal{A}|}
\sum_{a\in\mathcal{A}}
\frac{1}{|\mathcal{T}_a|}
\sum_{t\in\mathcal{T}_a} E_{t,a} +\lambda
\|\operatorname{normalize}(\boldsymbol{\theta})\|_2^2,
\end{aligned}
\label{eq:camera_robot_refinement}
\end{equation}
where $\mathcal{B}$ is the feasible set of corrections,
$\mathcal{A}$ contains the arm indices with valid masks in the
current view, and $\mathcal{T}_a$ contains the valid fitting frames
for arm $a$. The silhouette loss is
\begin{equation}
E_{t,a}
= 1-\operatorname{softIoU}(\widehat{M}_{t,a},M_{t,a}).
\end{equation}

For the head camera, we optimize three parameters,
\begin{equation}
\boldsymbol{\theta}_{\mathrm{head}}
=
(\Delta q_{\mathrm{yaw}},\,\Delta q_{\mathrm{pitch}},\,
 \Delta\phi_{\mathrm{roll}}).
\end{equation}
We bound the roll offset to $\pm10^\circ$. For yaw and pitch,
we bound the shared offsets so that the corrected head joint
angles remain within the robot's joint limits at every fitting
frame.
We set $\lambda=0$ for the head camera.

For each wrist camera, we optimize ten parameters,
\begin{equation}
\boldsymbol{\theta}_{\mathrm{wrist}}
=
(\Delta\boldsymbol{\omega},\,\Delta\mathbf{t},\,
 \boldsymbol{\alpha},\,\Delta\mathbf{c})
\in\mathbb{R}^{10}.
\end{equation}
The mounting rotation correction $\Delta\boldsymbol{\omega}$ is
a three-dimensional axis--angle vector in the initial camera frame.
Its direction gives the rotation axis, and its norm gives the angle
in radians. The vector $\Delta\mathbf{t}$ corrects the mounting
translation. The two entries of $\boldsymbol{\alpha}$ adjust the
focal lengths in log space, and the two entries of $\Delta\mathbf{c}$
shift the principal point. 

We bound the components of these four
groups to $\pm0.26\,\mathrm{rad}$, $\pm8.01\,\mathrm{cm}$,
$\pm0.30$, and $\pm60$ pixels, respectively. For each wrist camera, we set $\lambda=10^{-6}$.
The operator $\operatorname{normalize}$ divides the components of
the four correction groups by $0.1\,\mathrm{rad}$,
$3\,\mathrm{cm}$, $0.15$, and $20$ pixels, respectively.
The resulting $\ell_2$ penalty keeps the wrist calibration close
to its initial estimate.

\noindent\textbf{Training Details.}
We follow the training schedule of~\cite{liao2026genie,zhao2026real2edit2real}
and fine-tune all parameters of the GE-Sim-2B video generation
transformer for 12,000 steps. We use 8-bit AdamW with a learning rate
of $10^{-4}$, 1,000 warmup steps, and a global batch size of 16.
Our training set comprises 13,561 episodes across 147 tasks.
Each sample contains 25 target frames and four memory frames from
synchronized head, left-wrist, and right-wrist views at a resolution
of $384\times512$. We drop the memory with probability 0.5
during training.

\begin{figure*}[t]
    \centering
    \vspace{-15pt}
    \includegraphics[width=0.95\textwidth]{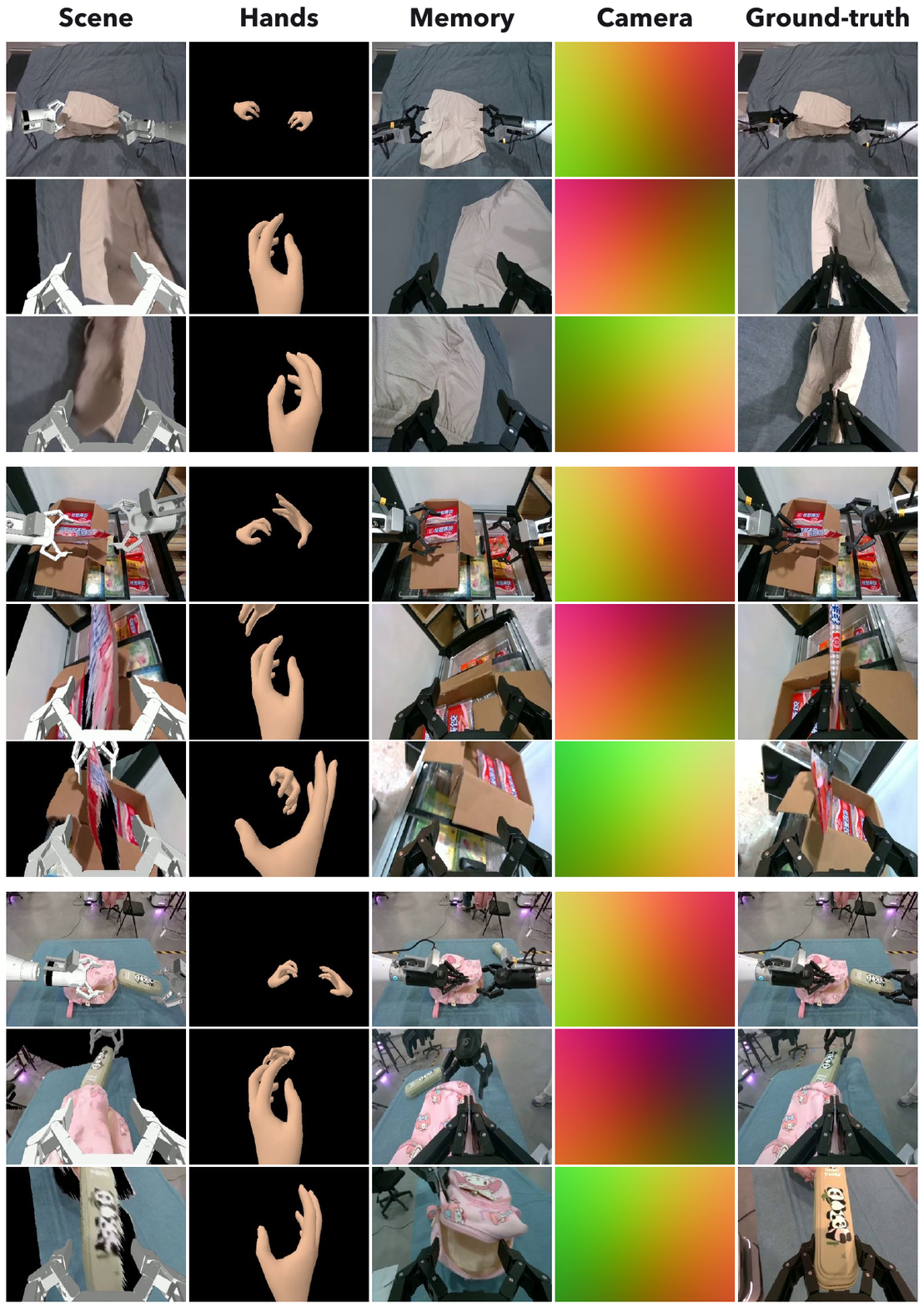}
    \vspace{-15pt}
    \caption{Representative training samples for multi-view video generation.}
    \label{fig:training_data}
\end{figure*}

\noindent\textbf{Training Data Visualization.}
Figure~\ref{fig:training_data} presents representative training samples
used for our multi-view video generation model.

\subsection{End-to-End Demonstration Synthesis}
\noindent\textbf{Hand Pose Refinement.}
We initialize the hand poses using HaWoR~\cite{zhang2025hawor}
with the calibrated focal length of the rectified left camera.
To reduce the remaining depth ambiguity, we refine the
MANO~\cite{MANO:SIGGRAPHASIA:2017} meshes using hand segmentation
masks and metric depth. Let $\boldsymbol{\theta}$ collect the
per-frame hand rotations, translations, and finger joint angles,
and let $\boldsymbol{\beta}$ denote the MANO shape vector shared
across both hands and all frames. For each hand, let $\mathcal{T}$
be the set of valid fitting frames. We minimize
\begin{equation}
\begin{aligned}
\min_{\boldsymbol{\theta},\,\boldsymbol{\beta}}\quad
&\frac{1}{|\mathcal{T}|}\sum_{t\in\mathcal{T}}
  \bigl(E_m^t+\lambda_d E_d^t\bigr)+\lambda_s E_s ,
\end{aligned}
\label{eq:hand_refinement}
\end{equation}
averaging the objective across the observed hands.

Let $\widehat M_t(p)$ and $M_t(p)$ denote the rendered and observed
hand mask values at pixel $p$. The mask loss measures the fraction
of the rendered silhouette that falls outside the observed mask:
\begin{equation}
E_m^t =
\frac{\sum_p \widehat M_t(p)\bigl(1-M_t(p)\bigr)}
     {\sum_p \widehat M_t(p)}.
\end{equation}

We back-project valid metric depths within the filtered hand mask
to obtain a point cloud $\mathcal{P}_t$. We select $N_t$ visible
points on the initial hand mesh and track their positions
$\mathbf{x}_{t,i}$ as the mesh deforms. The depth loss is
\begin{equation}
E_d^t =
\frac{1}{N_t}\sum_{i=1}^{N_t}\operatorname{dist}(\mathbf{x}_{t,i},\mathcal{P}_t),
\end{equation}
where $\operatorname{dist}$ denotes the Euclidean distance to
the nearest point in $\mathcal{P}_t$.

For temporal regularization, let $a_{t,j}$ denote the magnitude
of the world-space acceleration of joint $j$, normalized by
$1\,\mathrm{m/s^2}$. We compute accelerations from consecutive
joint positions and their timestamps. The smoothness loss is
\begin{equation}
E_s =
\frac{1}{|\mathcal{T}_s|J}
\sum_{t\in\mathcal{T}_s}\sum_{j=0}^{J-1}
a_{t,j}
+
\frac{\lambda_w}{|\mathcal{T}_s|}
\sum_{t\in\mathcal{T}_s} a_{t,0}^2,
\end{equation}
where $\mathcal{T}_s$ contains the centers of valid frame triples,
$J$ is the number of hand joints, and joint $0$ denotes the wrist.

We set $\lambda_d=1000$,
$\lambda_s=10^{-3}$, $\lambda_w=0.05$, and $N_t=2048$. We first refine the global rotation
and translation for 20 epochs, then jointly refine pose and shape
for 100 epochs.

\noindent\textbf{Tabletop Axes Alignment.}
Egocentric demonstrations are collected at varying head heights
and orientations. We align the human and robot head-camera
coordinate systems using the tabletop as a reference.

We implement RANSAC~\cite{fischler1981random} to fit a plane to the tabletop point cloud reconstructed from
the first RGB-D frame and segmented by SAM-3~\cite{carion2026sam}. Let $(\mathbf{n}_h,d_h)$ and
$(\mathbf{n}_r,d_r)$ denote the tabletop planes in the human
and robot head-camera frames, respectively,
\begin{equation}
\mathbf{n}_h^\top\mathbf{x}_h+d_h=0,
\qquad
\mathbf{n}_r^\top\mathbf{x}_r+d_r=0.
\end{equation}
The normals have unit length and point toward their respective
cameras, so $d_h,d_r>0$ are the perpendicular camera-to-table
distances.

We compute the minimum-angle rotation $R$ that aligns the
tabletop normals and a translation $\mathbf{t}$ that aligns
the camera-to-table distances,
\begin{equation}
R\mathbf{n}_h=\mathbf{n}_r,
\qquad
\mathbf{t}=(d_h-d_r)\mathbf{n}_r.
\end{equation}
The resulting human-to-robot head transform is
\begin{equation}
T_{h\rightarrow r}
=
\begin{bmatrix}
R & \mathbf{t}\\
\mathbf{0}^\top & 1
\end{bmatrix},
\qquad
\mathbf{x}_r=R\mathbf{x}_h+\mathbf{t}.
\end{equation}
We apply the same rigid transform to the reconstructed scene
and hand poses. 

\noindent\textbf{Trajectory Optimization.}
The joint trajectories obtained from cuRobo inverse
kinematics~\cite{sundaralingam2023curobo} may exhibit abrupt changes.
We optimize each arm trajectory to improve smoothness while
preserving the target end-effector poses.

Let $\mathbf{q}_t\in\mathbb{R}^{J}$ denote the optimized joint
angles at timestamp $\tau_t$, where $J=7$ is the number of joints
in each arm. For a trajectory of $T$ frames, we compute joint
velocities and accelerations as
\begin{equation}
\begin{aligned}
\mathbf{v}_t
&=\frac{\mathbf{q}_{t+1}-\mathbf{q}_t}
        {\tau_{t+1}-\tau_t},\\
\mathbf{a}_t
&=\frac{\mathbf{v}_{t+1}-\mathbf{v}_t}
        {\tau_{t+2}-\tau_{t+1}}.
\end{aligned}
\end{equation}
Let $\mathbf{p}(\mathbf{q}_t)$ and $R(\mathbf{q}_t)$ denote the
end-effector position and rotation obtained by forward kinematics,
and let $\mathbf{p}_t^*$ and $R_t^*$ denote their targets.
We define normalized pose errors
\begin{equation}
\begin{aligned}
\mathbf{e}_p^t
&=\frac{\mathbf{p}(\mathbf{q}_t)-\mathbf{p}_t^*}{\epsilon_p},\\
\mathbf{e}_R^t
&=\frac{R(\mathbf{q}_t)-R_t^*}
        {\sqrt{8}\sin(\epsilon_R/2)}.
\end{aligned}
\end{equation}
The acceleration, velocity, and pose losses are
\begin{equation}
\begin{aligned}
E_a &= \frac{1}{TJ}\sum_t\|\mathbf{a}_t\|_2^2,\\
E_v &= \frac{1}{TJ}\sum_t\|\mathbf{v}_t\|_2^2,\\
E_p &= \frac{1}{T}\sum_t
\bigl(\|\mathbf{e}_p^t\|_2^2+\|\mathbf{e}_R^t\|_F^2\bigr),
\end{aligned}
\end{equation}
where each sum includes only the valid time indices for its term.
We optimize
\begin{equation}
\begin{aligned}
\min_{\mathbf{q}_1,\ldots,\mathbf{q}_T}\quad
&\lambda_a E_a+\lambda_v E_v+\lambda_p E_p\\
\text{s.t.}\quad
&\mathbf{q}_{\min}\preceq\mathbf{q}_t
 \preceq\mathbf{q}_{\max},\\
&|\mathbf{v}_t|\preceq v_{\max}\mathbf{1},\\
&\|\mathbf{e}_p^t\|_2\leq1,\qquad
 \|\mathbf{e}_R^t\|_F\leq1.
\end{aligned}
\label{eq:trajectory_optimization}
\end{equation}
Here, $\mathbf{q}_{\min}$ and $\mathbf{q}_{\max}$ are the joint
limits, $v_{\max}$ bounds the joint velocities, and $\epsilon_p$
and $\epsilon_R$ specify the position and rotation tolerances.
We keep the prescribed initial states fixed and set
$\epsilon_p=0.0199\,\mathrm{m}$, $\epsilon_R=9.95^\circ$,
$v_{\max}=1.99\,\mathrm{rad/s}$, $\lambda_a=1/2500$,
$\lambda_v=0.02/v_{\max}^2$, and $\lambda_p=0.01$.
We solve the nonlinear problem through successive convex
approximations using Clarabel~\cite{goulart2026clarabel}.

\noindent\textbf{Collision Adjustment.} Due to the difference between human fingers and robot grippers, directly mapping human hand poses to robot wrist poses can lead to collision with the table. To alleviate the problem, first, we construct a tabletop collision model from the reconstructed RGB-D point cloud and detect intersections with the robot mesh.
Then, we adjust the wrist joint to rotate the gripper away
from the tabletop. For remaining collisions, we shift the
end-effector targets along the tabletop normal and optimize
the joint trajectories while enforcing joint and velocity limits.
Finally, we densely sample the adjusted trajectories and verify
that both the arms and grippers maintain the required tabletop
clearance.

\noindent\textbf{Head Stabilization.}
To construct conditions for multi-view video generation from a
fixed robot head viewpoint, we stabilize the egocentric observations
by reprojecting the RGB-D frames into the fixed robot head camera
using the aligned human-to-robot head transform.

\noindent\textbf{Gripper State to Action.}
We independently convert the estimated widths of each gripper into
binary commands ($0$ for open and $1$ for close).
We apply a five-frame median filter and detect sustained width
changes relative to a preceding stable plateau, and confirm an opening or closing event when the width deviates
from the plateau by at least $10\,\mathrm{mm}$ for $0.4$ seconds.
The gripper command is finally set to begin at the last plateau frame
before the motion starts.

\section{Experiment Details}

\subsection{Workspace and Objects}

\begin{figure*}[t]
    \centering
    \begin{minipage}[t]{0.49\textwidth}
        \centering
        \includegraphics[width=\linewidth]{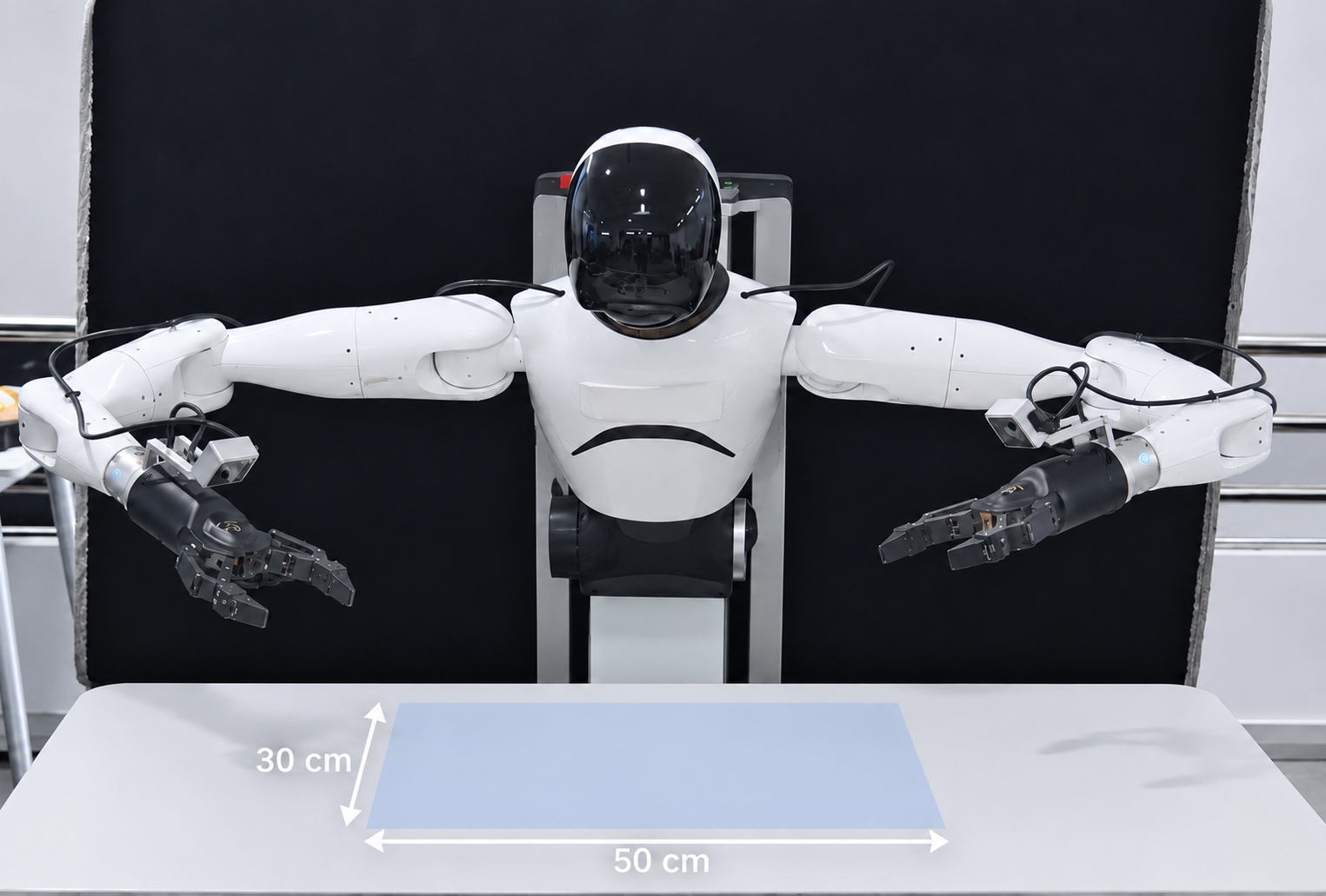}
        \caption{Shared tabletop workspace for the five real-robot manipulation tasks.}
        \label{fig:workspace}
    \end{minipage}
    \hfill
    \begin{minipage}[t]{0.49\textwidth}
        \centering
        \includegraphics[width=\linewidth]{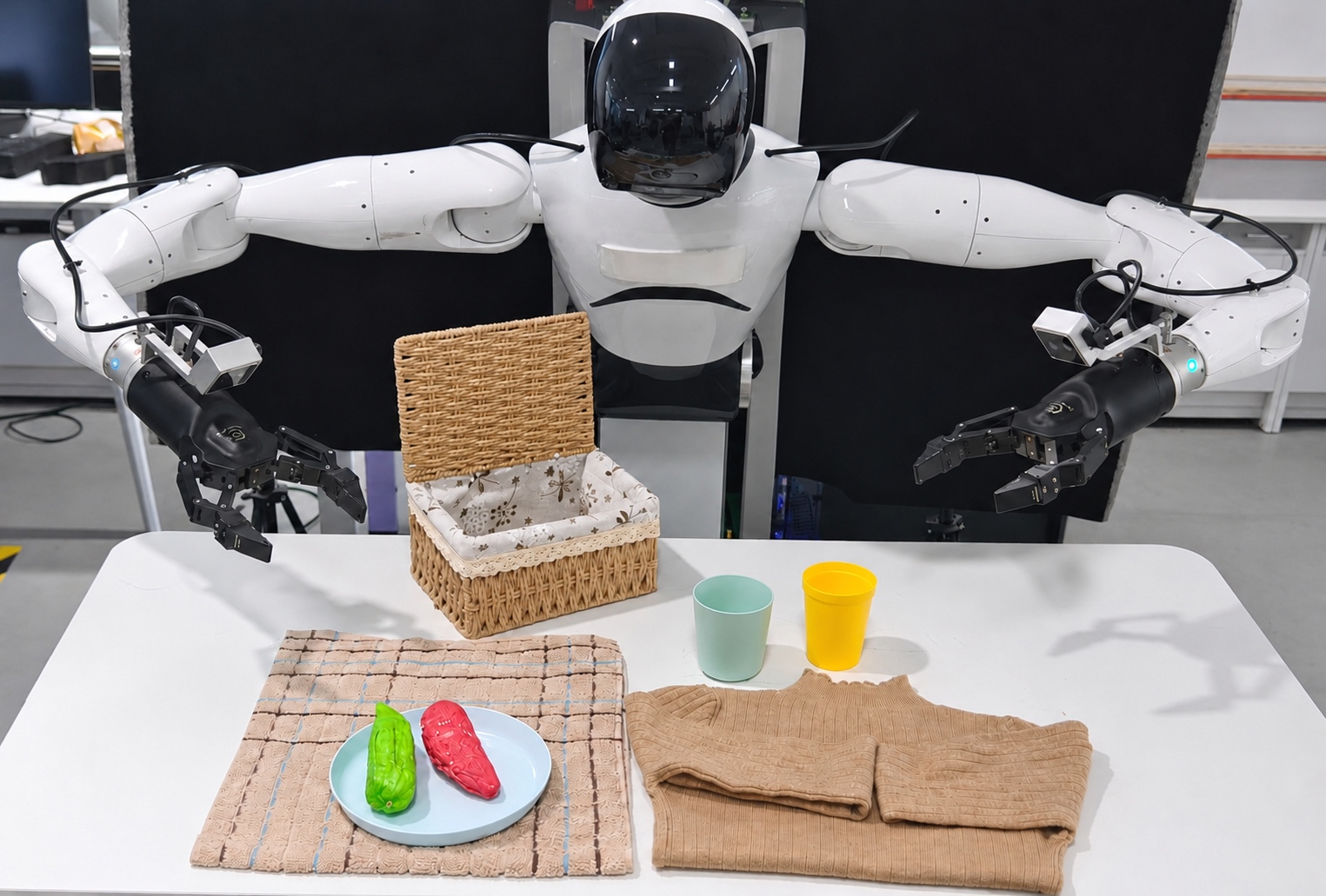}
        \caption{Objects used in the real-robot manipulation experiments.}
        \label{fig:object}
    \end{minipage}

    \medskip

    \begin{minipage}{\textwidth}
        \centering
        \includegraphics[width=\linewidth]{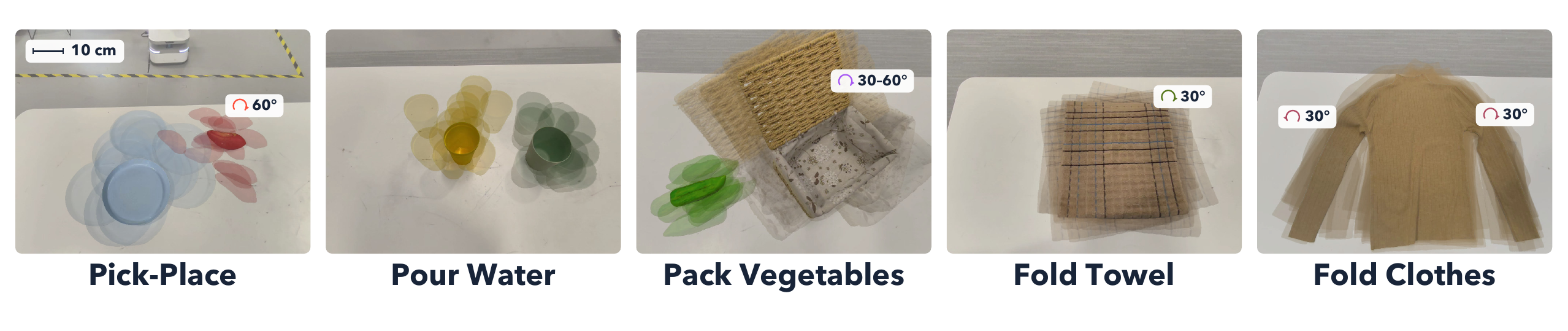}
        \caption{Visualization of the spatial distribution. We show 15 representative demonstrations selected from the full set of 50 for each task.}
        \label{fig:distribution}
    \end{minipage}
\end{figure*}

\noindent\textbf{Workspace.}
Figure~\ref{fig:workspace} shows the tabletop workspace shared by all
five real-robot manipulation tasks.

\noindent\textbf{Objects.}
Figure~\ref{fig:object} shows the objects used in the real-robot
experiments. The video model's training data exclude all environments
used for real-world evaluation.

\subsection{Spatial Distribution}

Figure~\ref{fig:distribution} illustrates the spatial distribution of object placements across our human demonstrations. All tasks share the tabletop workspace described in the main paper. Across demonstrations, the sweet potato in Pick-Place spans approximately \(60^\circ\) in yaw, and the basket lid in Pack Vegetables opens to angles between \(30^\circ\) and \(60^\circ\) relative to its closed position. The towel in Fold Towel spans approximately \(30^\circ\) in yaw, and each sleeve in Fold Clothes spans approximately \(30^\circ\) relative to the garment body.

\subsection{Baseline Implementation Details}

We implement Phantom~\cite{lepert2025phantom} and
Ego2Robot~\cite{wang2026ego2robot} within our demonstration
synthesis pipeline using the same 50 egocentric human
demonstrations per task. Each baseline uses its own human-to-robot retargeting algorithm and single-view observation
rendering procedure in place of our robot action retargeting
and multi-view video generation, respectively. Since Ego2Robot does not provide binary gripper action annotations, we manually annotate gripper open/close states by inspecting uniformly sampled keyframes from the generated observations. All remaining
pipeline components are shared across methods. We apply
identical sample selection criteria and use the same
downstream policy training and evaluation settings for
all methods.

\section{Additional Experiments}

\input{tables/moreablation}

\subsection{Additional Ablation Study}
To evaluate the contributions of multi-view video generation
and gripper state estimation, we replace each component
with a simpler alternative. For multi-view video generation, we first project the
head-view observations into wrist views, then render
robot arm and gripper meshes onto all views using our
generated gripper states, following
Phantom~\cite{lepert2025phantom}.
We subsequently employ
Minimax-Remover~\cite{zi2026minimax}
to inpaint the missing regions in the wrist views,
without using our multi-view video generation model. For gripper state estimation, we replace our predicted
gripper states and actions with those obtained from the
finger-to-gripper mapping of
Ego2Robot~\cite{wang2026ego2robot},
and our corresponding gripper action annotations.

\noindent\textbf{Effect of Multi-view Video Generation.} As shown in Table~\ref{tab:policyablation},
replacing our multi-view video generation model with
the simple inpainting baseline reduces the average success
rate from 57\% to 42\%, highlighting the importance
of accurate visual synthesis for robot demonstration generation.
Most tasks exhibit degraded performance under this
substitution, likely due to inaccuracies in synthesized
robot-object interactions.
An exception is Pack Vegetables, where artifacts introduced
by our multi-view generation model may negatively affect
policy performance, while the inpainting baseline preserves
the original projected observations and achieves better results.

\noindent\textbf{Effect of Gripper State Estimation.}
As shown in Table~\ref{tab:policyablation},
even with high-quality gripper action annotations,
replacing our video-guided gripper state estimation
with direct finger-to-gripper mapping reduces the
average success rate from 57\% to 43\%,
demonstrating the effectiveness of our approach.
The largest performance drops occur in Fold Towel
(from 65\% to 25\%) and Fold Clothes
(from 50\% to 20\%), where successful manipulation
is particularly sensitive to accurate gripper states
and interaction timing.

\subsection{Gripper State Analysis}

\input{tables/ego2robotcomparison}

\input{tables/handstylecomparison}

\noindent\textbf{Comparison with Ego2Robot.}
We compare the gripper states generated by our method
and Ego2Robot in Table~\ref{tab:gripper_state_difference}.
The results show that the discrepancy increases with
task complexity, suggesting that direct finger-to-gripper
mapping struggles to capture diverse hand-object interactions.
In contrast, our method learns to infer gripper states
through reverse-synthesized supervision, offering a
data-driven approach that can benefit from more diverse
training data.
These findings are consistent with the performance
degradation observed in folding tasks in
Table~\ref{tab:policyablation}.

\noindent\textbf{Comparison with 1/5 Grasp Styles.}
To quantitatively evaluate the impact of grasp style diversity
on gripper state estimation, we replace our video generation
model with a variant trained using only one-fifth of the
grasp styles, while keeping the same egocentric human
demonstrations and all other pipeline components unchanged.
We assess the validity of the generated gripper action
sequences using task-specific rules.
For example, Pick-Place requires the right gripper to
close and open exactly once, while Fold Clothes requires
both grippers to close and open twice.
We then compute the proportion of generated sequences
satisfying these rules.
As shown in Table~\ref{tab:one_fifth_gripper_pass_rate},
reducing grasp style diversity decreases the validity
of generated gripper action sequences, particularly
for contact-sensitive tasks such as folding.

\input{tables/timeanalysis}

\input{tables/costanalysis}

\subsection{Time and Cost Analysis}

\noindent\textbf{Time Analysis.}
Table~\ref{tab:processing_time} reports the amortized processing time for generating a 20-second, 30-fps robot demonstration using eight data-parallel NVIDIA A100 GPUs. Video generation dominates the pipeline (65.1\%), with multi-view synthesis alone accounting for 50.7\%. The dominant overhead can potentially be reduced through advances in generative model acceleration, such as autoregressive distillation~\cite{huang2026roborender}. Moreover, since demonstrations are processed independently, the pipeline naturally supports data-parallel scaling.

\noindent\textbf{Cost Analysis.}
Table~\ref{tab:data_cost} estimates the cost per 20-second demonstration based on public labor wage benchmarks~\cite{bls2026wages} and GPU rental prices~\cite{hyperstack2026pricing}. At \$17.44 per labor hour and \$1.60 per A100 GPU-hour, \themethod costs approximately \$0.68 per demonstration, including human data collection and GPU inference. In contrast, commercial robot data collection services list prices of \$1--3 for episodes of comparable duration (15--30 seconds)~\cite{roboticscenter2026pricing}. Despite differences in pricing models, \themethod offers competitive costs while enabling versatile robot demonstration generation without requiring physical robot deployment.

\section{Additional Visualizations}

\begin{figure*}[t]
    \centering
    \vspace{-15pt}
    \includegraphics[width=0.95\textwidth]{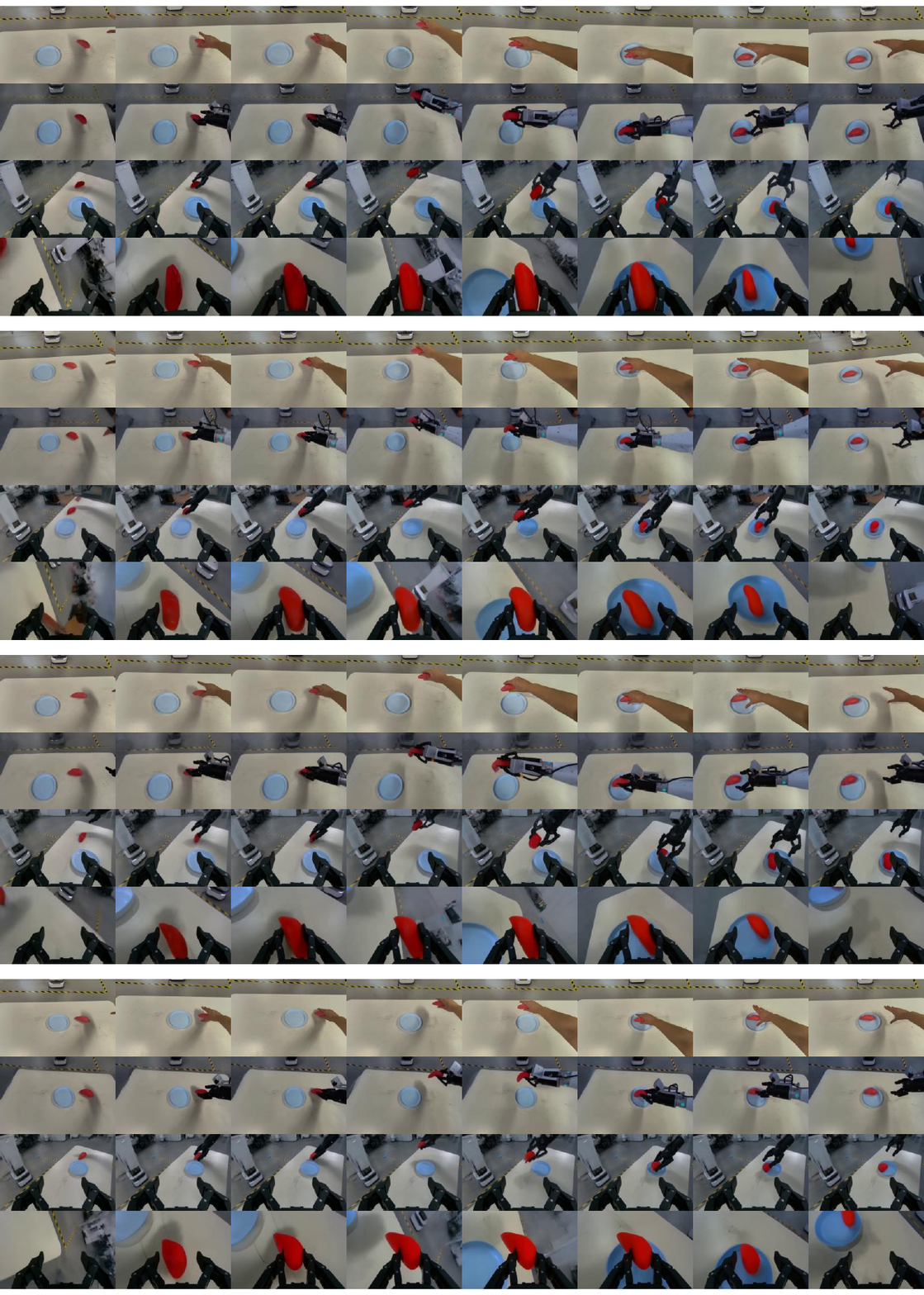}
    \vspace{-15pt}
    \caption{Visualization of generated videos on Pick-Place.}
    \label{fig:supp_vis_pickplace}
\end{figure*}

\begin{figure*}[t]
    \centering
    \vspace{-15pt}
    \includegraphics[width=0.95\textwidth]{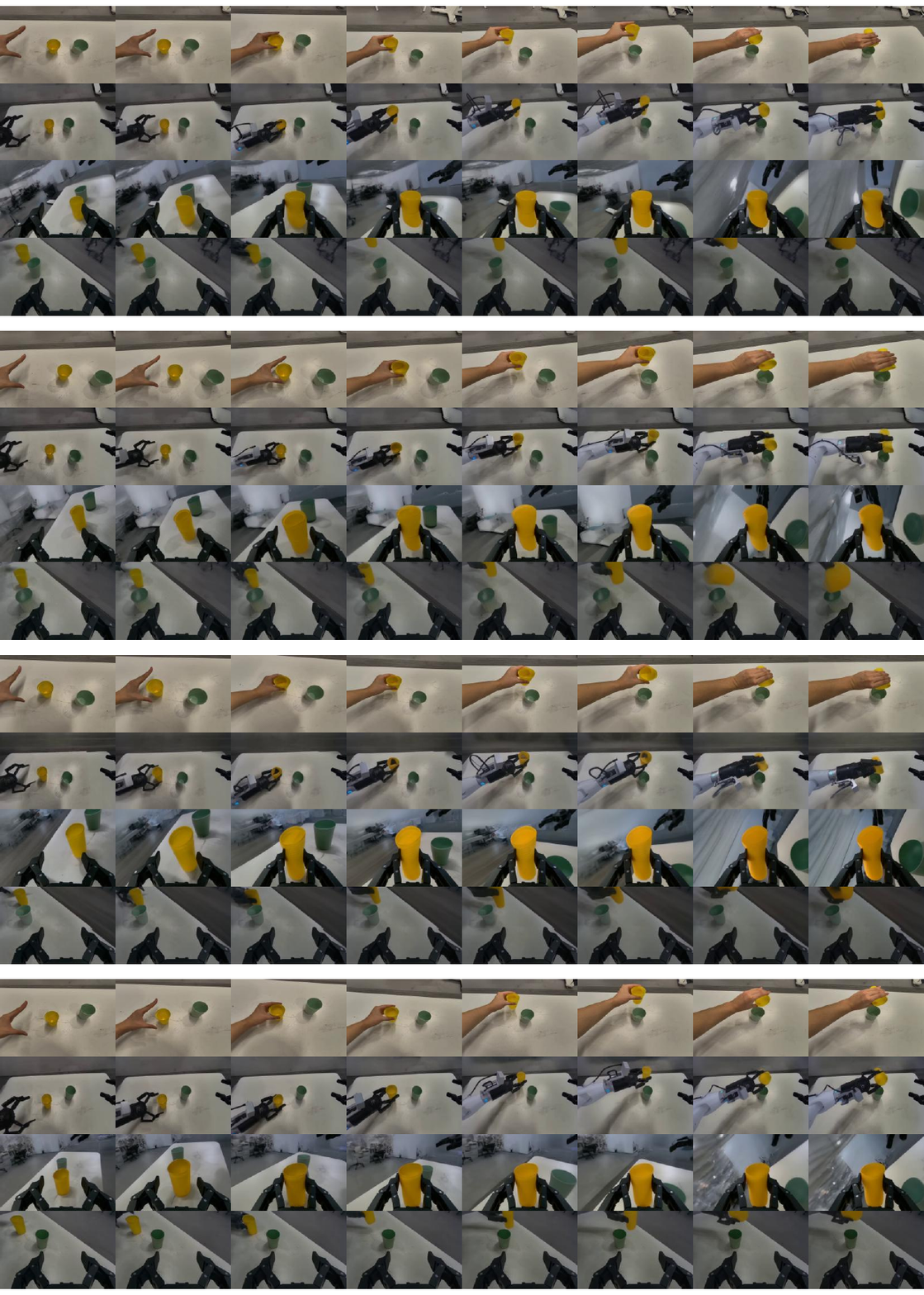}
    \vspace{-15pt}
    \caption{Visualization of generated videos on Pour Water.}
    \label{fig:supp_vis_pourwater}
\end{figure*}

\begin{figure*}[t]
    \centering
    \vspace{-15pt}
    \includegraphics[width=0.95\textwidth]{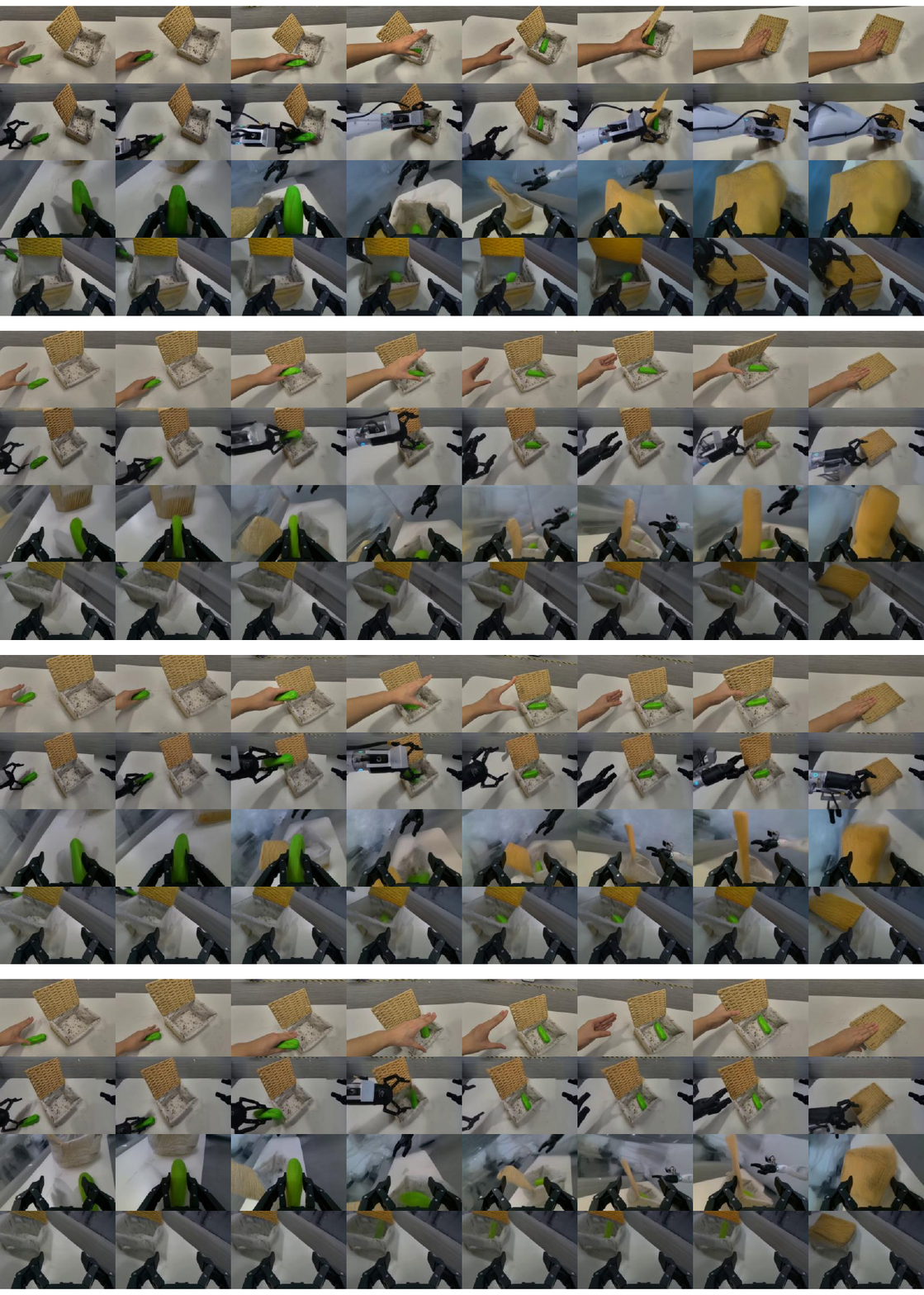}
    \vspace{-15pt}
    \caption{Visualization of generated videos on Pack Vegetables.}
    \label{fig:supp_vis_packvegetables}
\end{figure*}

\begin{figure*}[t]
    \centering
    \vspace{-15pt}
    \includegraphics[width=0.95\textwidth]{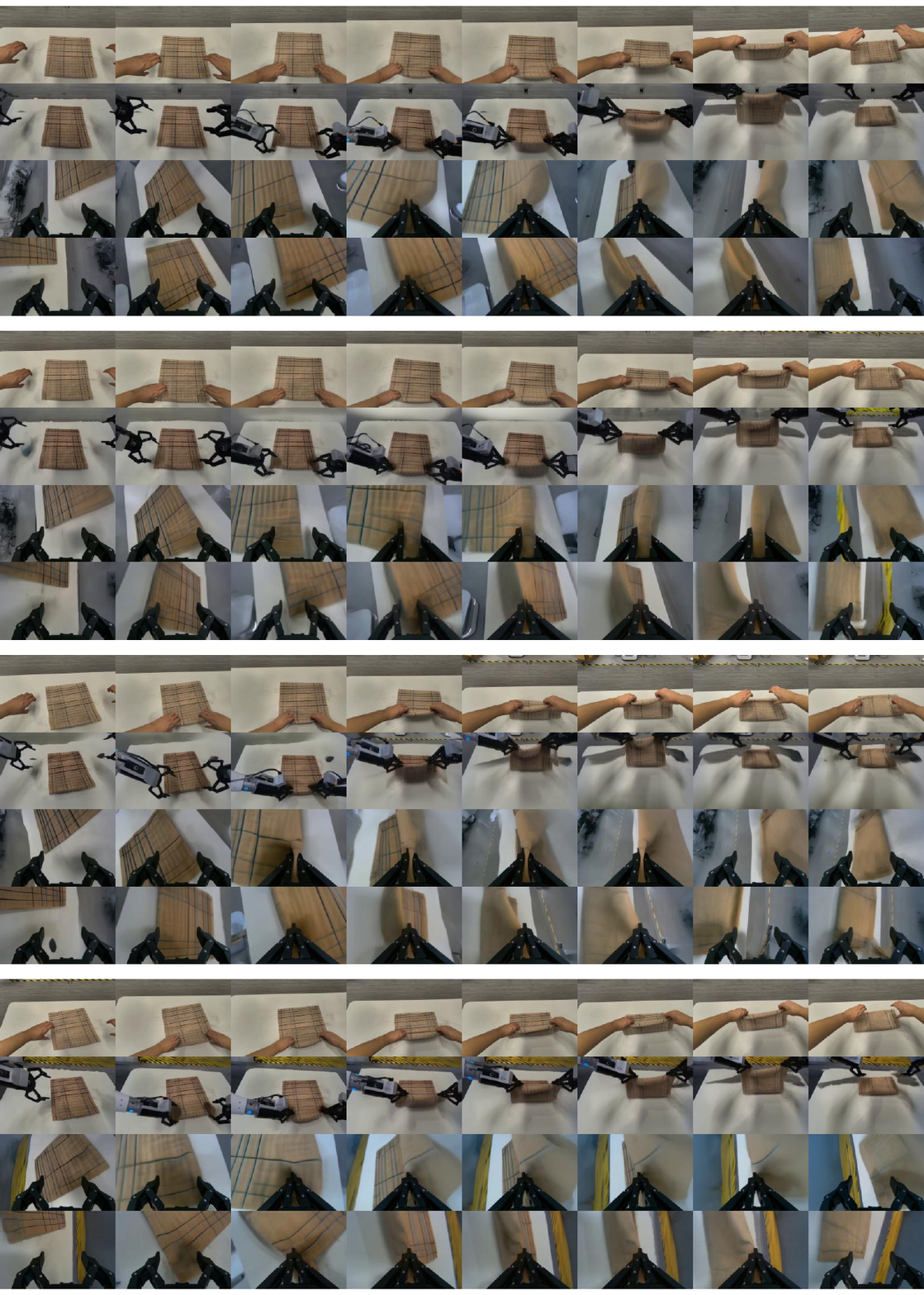}
    \vspace{-15pt}
    \caption{Visualization of generated videos on Fold Towel.}
    \label{fig:supp_vis_foldtowel}
\end{figure*}

\begin{figure*}[t]
    \centering
    \vspace{-15pt}
    \includegraphics[width=0.95\textwidth]{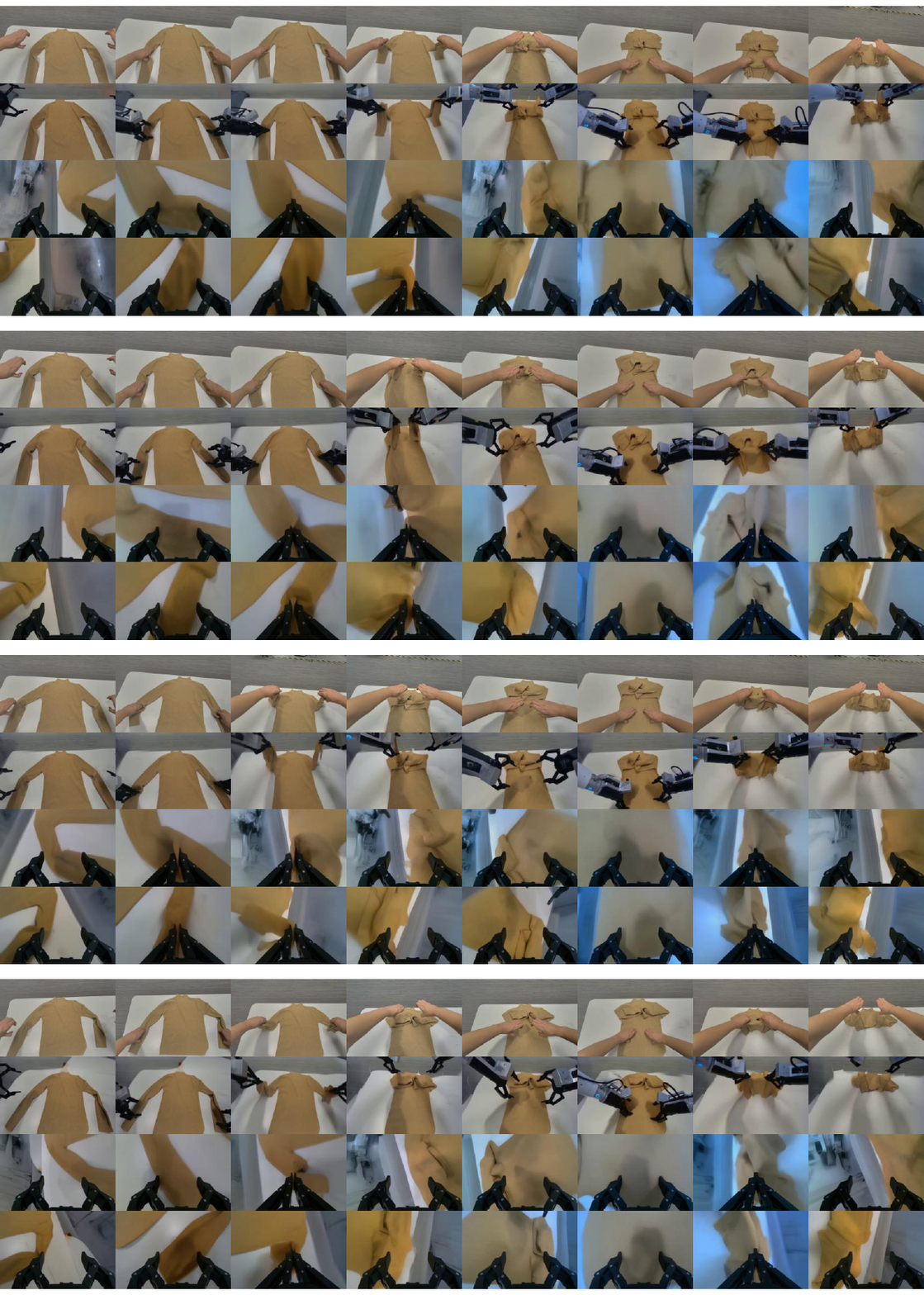}
    \vspace{-15pt}
    \caption{Visualization of generated videos on Fold Clothes.}
    \label{fig:supp_vis_foldclothes}
\end{figure*}

We demonstrate more visualizations of the five real-world manipulation tasks in Figure~\ref{fig:supp_vis_pickplace}, \ref{fig:supp_vis_pourwater}, \ref{fig:supp_vis_packvegetables}, \ref{fig:supp_vis_foldtowel}, \ref{fig:supp_vis_foldclothes}.

\section{Limitations and Discussions}

Although \themethod demonstrates the ability to convert
egocentric human data into diverse multi-view robot demonstrations,
several limitations remain.

\noindent\textbf{Human-to-Robot Action Gap.}
Our framework focuses on converting egocentric human demonstrations into actions for gripper-based robots rather than dexterous robotic hands. This limits its applicability to tasks requiring fine-grained finger manipulation, such as tying shoelaces and handling
small screws. Moreover, our reverse synthesis covers only a subset of possible gripper-to-hand configurations, potentially limiting the diversity of synthesized human hand motions.

\noindent\textbf{Occlusion in Egocentric Views.}
Our multi-view video generation relies on reconstructed 4D point clouds as intermediate representations, which may contain substantial missing regions due to occlusions in egocentric observations. Although our generative model can inpaint these regions, its ability to recover unseen content is constrained by the scale and diversity of the training data. Scaling up cross-view video generation training may help alleviate this limitation.

\noindent\textbf{Reliance on Foundation Models.}
\themethod builds upon pretrained vision foundation models for intermediate perception tasks, introducing the risk of error propagation throughout the synthesis pipeline. Nevertheless, its modular design allows individual components to benefit from future advances in foundation models without modifying the overall framework.

%% file: tables/methodcomparison.tex
\twocolumn[{
    \renewcommand\twocolumn[1][]{#1}
    \maketitlesupplementary
    \begin{center}
        \normalsize
        \renewcommand\tabcolsep{10pt}
        \resizebox{\textwidth}{!}{
        \begin{tabular}{l|cccc}
        \toprule
        \textbf{Method}
        & \textbf{No 3D Asset}
        & \textbf{Wrist-View Gen.}
        & \textbf{Multi-View Gen.}
        & \textbf{Scene Editing} \\
        \midrule
        \multicolumn{5}{l}{\textit{Scene Rendering \& Simulation Methods}} \\
        \midrule
        WARPED~\cite{freeman2026warped}
        & \ding{55} & \ding{51} & \ding{55} & \ding{51} \\
        EgoEngine~\cite{liu2026egoengine}
        & \ding{55} & \ding{55} & \ding{55} & \ding{55} \\
        Real2Render2Real~\cite{yu2025real2render2real}
        & \ding{55} & \ding{55} & \ding{55} & \ding{51} \\
        Real2Sem2Real~\cite{strong2026real2sem2real}
        & \ding{55} & \ding{55} & \ding{55} & \ding{51} \\
        \midrule
        \multicolumn{5}{l}{\textit{Video Editing \& Generation Methods}} \\
        \midrule
        H2R~\cite{li2025h2r}
        & \ding{51} & \ding{55} & \ding{55} & \ding{55} \\
        Phantom~\cite{lepert2025phantom}
        & \ding{51} & \ding{55} & \ding{55} & \ding{55} \\
        Masquerade~\cite{lepert2026masquerade}
        & \ding{51} & \ding{55} & \ding{55} & \ding{55} \\
        H2R-Grounder~\cite{ci2025h2r}
        & \ding{51} & \ding{55} & \ding{55} & \ding{55} \\
        MimicDreamer~\cite{li2025mimicdreamer}
        & \ding{51} & \ding{55} & \ding{55} & \ding{55} \\
        Ego2Robot~\cite{wang2026ego2robot}
        & \ding{51} & \ding{55} & \ding{55} & \ding{55} \\
        RwoR~\cite{heng2025rwor}
        & \ding{51} & \ding{51} & \ding{55} & \ding{55} \\
        \midrule
        \textbf{EgoMirror (Ours)}
        & \ding{51} & \ding{51} & \ding{51} & \ding{51} \\
        \bottomrule
        \end{tabular}
        }
        \captionof{table}{
            Comparison with prior human-to-robot demonstration
            conversion methods.
            No 3D Asset indicates that no 3D scene or object assets are required.
            Wrist Gen. and Multi-View Gen. denote generation of wrist views and of simultaneous multiple views.
            Scene Editing refers to augmenting
            demonstrations through spatial or texture
            editing of the original scene.
        }
        \label{tab:supp_compare}
    \end{center}
}]

%% file: tables/moreablation.tex
\begin{table*}[t]
\setlength{\tabcolsep}{4pt}
\centering
\begin{tabular}{l|ccccc|c}
\toprule
\textbf{Method}
& \textbf{Pick-Place}
& \textbf{Pour Water}
& \textbf{Pack Vegetables}
& \textbf{Fold Towel}
& \textbf{Fold Clothes}
& \textbf{Total} \\
\midrule
Phantom~\cite{lepert2025phantom}
& 0 / 20 & 2 / 20 & 0 / 20 & 1 / 20 & 0 / 20 & 3\% \\
Ego2Robot~\cite{wang2026ego2robot}
& 5 / 20 & 5 / 20 & 3 / 20 & 2 / 20 & 0 / 20 & 15\% \\
\midrule
Real (Single-view)
& 4 / 20 & 4 / 20 & 8 / 20 & 4 / 20 & 6 / 20 & 26\% \\
Ours (Single-view)
& 11 / 20 & 7 / 20 & 3 / 20 & 4 / 20 & 9 / 20 & 34\% \\
\midrule
Real (Multi-view)
& 12 / 20 & 10 / 20 & 11 / 20 & 9 / 20 & 10 / 20 & 52\% \\
Ours (Minimax-Remover Inpainting)
& 9 / 20 & 3 / 20 & 15 / 20 & 7 / 20 & 8 / 20 & 42\% \\
Ours (Ego2Robot Gripper Mapping)
& 14 / 20 & 11 / 20 & 9 / 20 & 5 / 20 & 4 / 20 & 43\% \\
Ours (Multi-view)
& 14 / 20 & 10 / 20 & 10 / 20 & 13 / 20 & 10 / 20 & 57\% \\
\bottomrule
\end{tabular}
\caption{
Ablation study of multi-view video generation and gripper state estimation. For comparison, we implement independent view inpainting and finger-to-gripper mapping as alternatives. We report policy success rates on five real-world robotic manipulation tasks, with 20 evaluation trials per task.
}
\vspace{-10pt}
\label{tab:policyablation}
\end{table*}

%% file: tables/ego2robotcomparison.tex
\begin{table}[t]
    \centering
    \setlength{\tabcolsep}{5pt}
    \begin{tabular}{lrr}
        \toprule
        \textbf{Task}
        & \shortstack{\textbf{Mean absolute}\\\textbf{difference (mm)}}
        & \shortstack{\textbf{Difference}\\\textbf{$>20$ mm (\%)}} \\
        \midrule
        Pick-Place      &  9.4 & 11.1 \\
        Pour Water      &  9.9 & 14.0 \\
        Pack Vegetables & 16.5 & 35.6 \\
        Fold Towel      & 25.6 & 68.0 \\
        Fold Clothes    & 29.1 & 65.8 \\
        \midrule
        Overall (Framewise) & 20.5 & 45.7 \\
        \bottomrule
    \end{tabular}
    \caption{Differences in estimated gripper opening between
    \themethod{} and Ego2Robot.}
    \label{tab:gripper_state_difference}
\end{table}

%% file: tables/handstylecomparison.tex
\begin{table}[t]
\centering
\begin{tabular}{lrr}
\toprule
\textbf{Task} & \textbf{Passed / Total} & \textbf{Pass Rate} \\
\midrule
Pick-Place       & 50 / 50   & 100.00\% \\
Pour Water       & 49 / 50   & 98.00\%  \\
Pack Vegetables  & 42 / 50   & 84.00\%  \\
Fold Towel       & 0 / 50    & 0.00\%   \\
Fold Clothes     & 5 / 50    & 10.00\%  \\
\midrule
Overall          & 146 / 250 & 58.40\%  \\
\bottomrule
\end{tabular}
\caption{Gripper action sequence pass rates with one-fifth of the grasp styles. A sequence passes if its estimated gripper open/close transitions satisfy the predefined task-specific rules.}
\label{tab:one_fifth_gripper_pass_rate}
\end{table}

%% file: tables/timeanalysis.tex
\begin{table}[t]
\centering
\small
\begin{tabular}{lr}
\toprule
\textbf{Task} & \textbf{Amortized Time (s)} \\
\midrule
\textbf{1. 4D Perception} & \\
1.1 Video Preprocessing & 0.26 \\
1.2 Stereo Depth Estimation & 10.10 \\
1.3 Forearm Segmentation & 8.79 \\
1.4 Scene Depth Completion & 3.88 \\
1.5 Hand Pose Estimation & 3.41 \\
1.6 Hand Refinement & 16.82 \\
\midrule
\textbf{2. Robot Action Retargeting} & \\
2.1 Axes Alignment & 0.50 \\
2.2 IK and Trajectory Optimization & 3.92 \\
2.3 Collision Adjustment & 2.91 \\
\midrule
\textbf{3. Video Generation} & \\
3.1 Head Stabilization & 1.34 \\
3.2 Hand Inpainting & 14.60 \\
3.3 Background Depth Completion & 4.25 \\
3.4 Multi-view Reprojection & 3.34 \\
3.5 Multi-view Video Generation & 82.80 \\
\midrule
\textbf{4. Gripper State Estimation} & \\
4.1 Gripper Segmentation & 6.28 \\
4.2 Template Matching & 0.07 \\
4.3 State and Action Estimation & 0.03 \\
\midrule
\textbf{Total} & \textbf{163.30 (2.72 min)} \\
\bottomrule
\end{tabular}
\caption{Amortized processing time per 20-second,
30-fps Pick-Place demonstration under
8$\times$ NVIDIA A100 80GB, assuming ideal linear scaling.}
\label{tab:processing_time}
\end{table}

%% file: tables/costanalysis.tex
\begin{table}[t]
\centering
\small
\setlength{\tabcolsep}{6pt}
\begin{tabular}{lcc}
\toprule
\textbf{Component} & \textbf{Unit Price} & \textbf{Cost (\$)} \\
\midrule
Human Data Collection
& \$17.44/h~\cite{bls2026wages} & 0.10 \\
A100 GPU Inference
& \$1.60/GPU/h~\cite{hyperstack2026pricing} & 0.58 \\
\textbf{\themethod (Ours)}
& -- & \textbf{0.68} \\
\midrule
Commercial Robot Data
& -- & 1.00--3.00~\cite{roboticscenter2026pricing} \\
\bottomrule
\end{tabular}
\caption{Estimated cost per 20-second Pick-Place
demonstration.}
\label{tab:data_cost}
\end{table}